# Symbolic Regression of Dynamic Network Models

*A Thesis* submitted to

Indian Institute of Science Education and Research, Pune

in partial fulfillment of the requirements for the

BS-MS Dual Degree Programme

by

Govind Gandhi

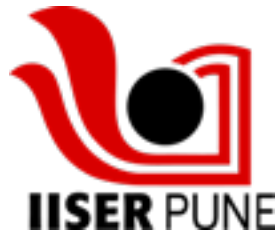

Indian Institute of Science Education and Research Pune

Dr. Homi Bhabha Road,

Pashan, Pune 411008, INDIA.

May, 2022

Supervisor: Dr. Camille Roth & Dr. Telmo Menezes

© Govind Gandhi (2022)

# Certificate

This is to certify that this dissertation entitled Symbolic Regression of Dynamic Network Models towards the partial fulfilment of the BS-MS dual degree programme at the Indian Institute of Science Education and Research, Pune represents study/work carried out by Govind Gandhi at Indian Institute of Science Education and Research under the supervision of Dr. Camille Roth & Dr. Telmo Menezes, Digital Humanities, CNRS, during the academic year 2021-2022.

<div style="text-align:right">Dr. Camille Roth & Dr. Telmo Menezes</div>

Committee:

Dr. Camille Roth & Dr. Telmo Menezes

Dr. Anindya Goswami

This thesis is dedicated to the number 42.

# Declaration

I hereby declare that the matter embodied in the report entitled Symbolic Regression of Dynamic Network Models are the results of the work carried out by me at the Indian Institute of Science Education and Research, Pune, under the supervision of Dr. Camille Roth & Dr. Telmo Menezes and the same has not been submitted elsewhere for any other degree.

<div style="text-align: right;">Govind Gandhi</div>

# Acknowledgments

I would like to express my deepest gratitude to my wonderful supervisors Dr. Camille Roth and Dr. Telmo Menezes. Without their mentorship, time and empathetic understanding during a turbulent year, this project would not have existed.

I am indebted to my friends for their compassionate[1] support and for making me realise that you can't walk that far if you're not walking together.

I thank my parents for setting me up and supporting me on a platform where I'm able to explore my interests and exercise my will.

I acknowledge assistance from the Scholarship for Higher Education (SHE), DST, Govt. of India and am extremely grateful to IISER Pune for making me the person I am today and for being an enriching sandbox - it doesn't hurt much if you try something and fall down here.

---

[1] Fun fact: Compassion literally means "to suffer together."

# Abstract

Growing interest in modelling complex systems from brains to societies to cities using networks has led to increased efforts to describe generative processes that explain those networks. Recent successes in machine learning have prompted the usage of evolutionary computation, especially genetic programming to evolve computer programs that effectively forage a multi-dimensional search space to iteratively find better solutions that explain network structure. Symbolic regression contributes to these approaches by replicating network morphologies using both structure and processes, all while not relying on the scientist's intuition or expertise. It distinguishes itself by introducing a novel formulation of a network generator and a parameter-free fitness function to evaluate the generated network and is found to consistently retrieve synthetically generated growth processes as well as simple, interpretable rules for a range of empirical networks. We extend this approach by modifying generator semantics to create and retrieve rules for time-varying networks. Lexicon to study networks created dynamically in multiple stages is introduced. The framework was improved using methods from the genetic programming toolkit (recombination) and computational improvements (using heuristic distance measures) and used to test the consistency and robustness of the upgrades to the semantics using synthetically generated networks. Using recombination was found to improve retrieval rate and fitness of the solutions. The framework was then used on three empirical datasets - subway networks of major cities, regions of street networks and semantic co-occurrence networks of literature in Artificial Intelligence to illustrate the possibility of obtaining interpretable, decentralised growth processes from complex networks.

# Contents

# List of Figures

# List of Tables

# Chapter 0

# Introduction

## 0.1 Motivation

We are deluged with systems of near-mystical complexity. A network/graph is a catalogue of a system's components, often called nodes/vertices and the direct interactions between them, called links/edges. The network paradigm has provided a successful framework to study the intricate patterns of relations among the constitutions of real-world complex systems. It has revealed that the dynamical behaviours observed in such systems, such as information spreading, diffusion, opinion formation, and synchronization, are quite often affected - and to some extent determined - by the **structure** of the underlying interaction network [1, 2, 3]. Therefore an understanding of the morphogenesis of the network and the processes that lead to its structure and eventually some control of its structural reconstruction is desirable.

The goal is to obtain certain processes or restrictions that are thought to be important in explaining network creation, such as transitivity, centrality, homophily, etc. Such a mechanism's role can be determined *a priori* by assuming its existence and checking if and how it contributes to the generation of a given network shape. Another way is assuming *a posteriori* and estimating the magnitude of influence of a mechanism during network evolution. In all circumstances, intuition is unavoidable: developing these models necessitates understanding which systems may be involved. However, the role of some processes, and more importantly, their combination, might seem counter-intuitive at times. In order to reduce this reliance on intuition, evolutionary algorithms were utilised recently to infer possible mechanisms from observed structures automatically. More specifically, network structure is utilised to devise edge formation processes, which are then used to precisely reconstruct the structure in an iterative process (more in section 1.1).

Template models based on fixed sets of possible actions were used in some of the early techniques (e.g., creating/rewiring an edge, tuning a node property, etc). Actions can be organised in a variety of ways, including as a fixed chart that resembles the typical structure of agent-based models [4], as a sequential list of variable size [5, 6]), or, more recently, as a matrix whose weights describe the relative contribution of each action [7]. The evolutionary processes used in all of these studies try to automatically fill the template model with actions and fit the necessary parameters to retrieve the network structure back. They use fitness functions to measure the similarity between the empirical network and the networks generated by the evolved model, as is customary in evolutionary programming. Often, classic structural attributes like degree distributions, motifs, centrality measures, etc., are used in fitness functions. Models are then iteratively improved as fitness values rise.

An original method for inferring arbitrarily complicated combinations of elementary processes, understood as rules, using the local information available in a network was proposed [8, 9]. This machine learning-based framework was able to provide interpretable rules; unlike many ML approaches which end up being black boxes; while achieving independence from the intuition required to supply explanatory processes for networks.

Ref. [9] has previously established a generic vocabulary that allows us to define network evolution as an iterative process based on the likelihood of an edge between two nodes appearing in the currently evolving network, understood as a function on node attributes. They used both structural such as distance and connectedness measures (like degree, betweenness, closeness centralities, etc) and non-structural node attributes like sequential identifiers. They were able to employ genetic programming approaches to evolve rules that were then used to produce network structures that were increasingly similar to the target network by representing these functions as trees. The purpose of this technique is to 'genetically' generate freeform symbolic expressions rather than fitting parameters linked with predefined symbolic expressions *i.e.* they automatically evolve realistic morphogenetic rules from a given instance of the target network by symbolically regressing it. This technique is inspired by Ref. [10], who extracted free-form scientific laws from experimental data. This technique was then first used on social networks [8], which led to a much more comprehensive framework in Ref. [9]. A notable achievement of this framework was the ability to discover the rules of an Erdos-Renyi or Barabasi-Albert generative process systematically and precisely from one of its stochastic realisations. There were also distinct, realistic, and compact formalisation of generators for a range of social, physical, and biological networks.

Their approach presented a kind of 'artificial scientist' suggesting plausible network models, replacing the intuition of a person modelling and suggesting a network generative process of a particular situation using their domain expertise. This approach allows for the discussion of networks in terms of their genotype (*i.e.*, generator equations) rather than phenotypes (i.e., a series of topological traits). Phenotypical traits serve as a foundation for evaluating the quality of structural reconstruction and, as a result, for creating fitness functions that are sensitive to specific topological properties Ref. [11]. They also provide a solid foundation for comparing networks via the fitness function. The choice of the attributes considered for it is by building over numerous studies that have tried to explain the role of different attributes in identifying network families. Some of these attributes investigated are triadic profiles [12], canonical analysis of various measures[11], block modelling [13], community structure [14], hierarchical structure [15] among others. Ref. [16] used evolutionary algorithms to symbolically regress formulas characterising the network based on its morphological features (phenotypes) of the network, for example, stating directly the width of various network classes as a function of the number of nodes, links, or eigenvalues of the adjacency matrix. Symbolic regression, on the other hand, allows for network comparison and categorization based on their plausible underlying morphogenesis rules − a genotypic categorization.

**Our objective** is to extend this method to dynamic networks, deal with the different complexities of carrying out such an extension and expectantly use them to display the involvement of underlying structural temporal patterns.

## 0.2   Goals

The objectives of this thesis are the following:

1. To theoretically describe and appraise the symbolic regression framework (`synth`) proposed in Ref. [9].

2. To propose and implement augmentations to `synth` that provide a way to capture generative rules for dynamic networks.

3. To test the modifications to `synth` *in-silico* and venture into using it on empirical networks.

Our primary focus is on the problem of obtaining rules from network structure and in effect, `synth`'s ability to recover rules from different kinds of information of the network and how we can interpret and make use of such information. Dealing with challenges like analysing the computational complexities of different parts of the algorithm, optimising the code or an exact analytical description of the rule generation process are interesting pursuits of their own accord and very much essential to make progress in our problem as well. However, they require their own theses and are out of our scope for this one.

## 0.3 Structure

This thesis consists of 5 chapters:

- Chapter 1 contextualises `synth` and places it within the existing literature and the state of art and explains the workings of the algorithm.

- Once an understanding of `synth` is established, the exploration of defining dynamic networks and extending the current framework to incorporate them is done in chapter 2. We test the consistency of the suggested improvements to the framework using *in-silico* tests.

- With a reliance on the framework established, chapter 3 presents some of the explorations done on real life data sets.

- We tie it all up and present our conclusions in chapter 4.

- The symbolic regression algorithm is complex and in appendix A some of the methods and techniques used or modified in this work is detailed. While these are no doubt essential to our work, we have placed it at the end so as to not detract from the main narrative of the thesis.

# Chapter 1

# Background

This chapter aims to lay an understanding of the important aspects of `synth` - its context, novelty, and its functioning. These are laid out in detail in the next three sections, respectively. At the end of the chapter some avenues of improvements are discussed and the direction of the thesis is laid out.

## 1.1  Where does `synth` fit in?

The problem of arriving at explanatory generative models for any given network is complex. There have been different kinds of efforts since the new millennium. The related state of the art could be organised according to two essential dichotomies [17]: the first one concerns the *target* of the modelling process and the second one to the *base* on which the modelling process is founded upon. In the former, the target of the model could be thought of as retrieving the evolutionary process or the network structure; and in the latter, the foundations of the modelling (the assumptions, essentially) can be fed via the process or the structure. This leads to a double dichotomy shown in Table 1.1. Where does our symbolic regression framework (`synth`) fit in this double dichotomy table?

| reconstructing / using | processes | structure |
|---|---|---|
| **processes** | Preferential Attachment, Link Prediction, Classifiers, Scoring methods, etc. | PA-based models, Rewiring models, Agent-based models, Random Walk based strategies, etc. |
| **structure** | ERGMs, Markov graphs, etc. | Subgraph sampling, Edge swaps, etc. |
| | **Symbolic Regression** | |

Table 1.1: The double dichotomy table of canonical network modelling approaches, set out to reconstruct either evolution processes or network structure, by relying either on evolution processes or network structure. Symbolic regression reconstructs the generative process and the structure using structure. - Modified from Table 1 of Ref. [17].

We can reconstruct the process, *i.e.* the mechanism through, which nodes/edges are added/deleted using processes. For instance, using the Preferential Attachment (PA) which is in itself a micro-level process that directs addition/deletion of new edges preferentially based on weights given by a certain kind of quantity like degree of a network, a geometric quantity for an embedded network, a centrality measure, etc. So if the connection probabilities of the process in question are replicated by the PA process based on a certain quantity (found through intuition or expertise or ML), then the evolutionary process and its rules is said to have been reconstructed using processes.

We can reconstruct the process using the structure of the network as well. For example,

the usage of Exponential Random Graph Models (ERGMs) assume that the network at hand is drawn from a distribution of networks. Now the objective is to find the parameters associated with different topological and/or non-topological attributes of the networks that maximise the probability of getting the given network from the distribution of networks. We lose the interpretability by reducing the contribution of different properties to scalar parameters but we have the ability to take into account a wide range of properties.

Moving to the other part of the double dichotomy table, we can once again construct structure of the network using processes as well as structure. Reconstructing structure using process is the most familiar modelling method in the statistical physicist's toolkit. Proposing PA-based models, agent-based models, link rewiring methods, (modified) random walking strategies all fall under this umbrella where we use these to direct network creation and try to replicate the properties of the target network like the degree distribution, connectivity measures, community structure, etc.

Construction of structure using structure entails the demonstration of structural constraints or attributes correlating to properties of the network in question. For instance, sampling subgraphs for connected components or degree structure to propose some sort of scaling; or identifying smaller units of network structure like dyads or cliques or simplices and inferring and verifying from their distribution the overall makeup of the structure. Amongst the four, this is the most arcane subsection owing to practical difficulties in sampling and limitations in bayesian methods of causal inference.

In all of the four sections, intuition was key, be it either to propose the involvement of a process or the attributes of network topology (centrality, homophily, transitivity, etc.) that shape the resulting network. The modus operandi of evolutionary algorithms fit well to remove this dependence on intuition. This is where `synth` comes in. **The symbolic regression framework jointly uses structure to reconstruct the process and the processes to reconstruct the structure.** When applied in the context of networks, the network structure is used to get at edge creation processes, and these processes are then used to discover, iteratively, other processes that can get progressively better in reconstructing the input network structure.

## 1.2  Novelty of the algorithm

The approach proposed in Ref. [9] employs a form of machine learning from evolutionary computation known as **genetic programming**. Evolutionary computation is a sort of natural selection-inspired search in which evolutionary pressure is used to steer a population of solutions to become increasingly better. Genetic programming is a class of evolutionary algorithms where solutions are computer programs. The iterative improvement of this population of solutions through mutation of the best equations as determined by some error metric (e.g. root mean squared error, or in this case, a more elaborate fitness function). In other words, genetic programming mimics natural selection in order to perform tasks including symbolic regression [18]. Ref. [5] is an early work applying this technique to networks. In our scenario, the population consists of network generative models, and the performance measure is how well a model's synthetic network approximates the real observable network.

Firstly it deals with a gap in current network science literature: There is no consistent and simple technique to express network generating processes in a formal fashion. They came up with the notion of the network generator as a computer program, which we'll refer to as generators. There is a set of potential edges that can be produced at any given time. If a generator allows you to prioritise one edge above the others, it is said to be completely specified. They design a stochastic selection process instead of attempting to describe a deterministic one, recognising that many of the generative processes that form networks include some inherent unpredictability.

They introduced a vocabulary of structural and non-structural node attributes that can be used to create a function that gives out probabilities allowing to pick an edge from a given sample set in the currently evolving network. Representing these functions as trees enabled them to apply genetic programming techniques to evolve rules which are then used to generate network morphologies increasingly similar to the target, empirical network. This is what we mean by "symbolic regression", whose goal is to genetically evolve free-form symbolic expressions rather than fitting parameters associated with fixed symbolic expressions. This strategy is inspired by [10] who extract free-form scientific laws from experimental data. The symbolic regression method was demonstrated to have the ability to exactly discover the laws of an Erdos-Renyi or Barabasi-Albert generative process from one of its stochastic realizations. It was then applied on social, physical and biological networks to obtain characteristic, concise, cogent laws [17]. On a side note, Ref. [19] demonstrates the efficacy of symbolic regression to get free-form equations from data, which is a much simpler version of the template of the problem we have at hand.

Secondly, there is the issue of measuring network similarity. They combine distributions of simple yet fundamental aspects of the network's identity like the degree, centrality measures, distance measures, triadic profiles. By comparing the normalised dissimilarity between the distributions of the synthetic and real-world networks we arrive at a measure of similarity that can be used across networks. This also makes it possible to discuss networks in terms of their plausible genotype (i.e., generator equations), rather than using their topological attributes (*i.e.* phenotypes). However, symbolic regression allows comparison based on the genotypes used to characterise the networks. Ref. [17] also discovered network families based on their genotypes (generator expressions) instead of their phenotypes (morphologies). Each network then has a generator expressed as a formula that explains its identity.

A succinct picture summarising the evolutionary algorithm described above (`synth`) is illustrated in Fig. 1.1.

## 1.2.1 The Network Generator

As mentioned in the previous section, we represent the generating mechanism that drives the creation and identity of a network as a *tree-based computer program*. A *tree* is an undirected network in which any two nodes are connected by exactly one path (equivalently, a connected acyclic undirected network). Here, we use a rooted tree in which an ordering is specified for the children nodes of each nodes. What does the generator represent using this tree data structure? The network generating mechanism is defined as a stochastic process where at each step, a new edge is added. Among (a subset of) the possible node pairs between which a new edge can be added, the way we decide between them is by assigning a weight/score to each of the randomly sampled candidate edges.

At each step of the network generation process, we randomly sample a subset of the candidate edges and assign weights according to the generator only to that subset and add an edge from this sample. Given a network with nodes belonging to set $V$ and having edges belonging to $E$ ($E \subseteq V \times V$). We consider the set difference between the set of all edges and the set of edges currently in the network *i.e.* $A = V \times V - E$. We introduce a predefined parameter called the sampling ratio $s_r \in (0, 1]$ and define a sample $S$ such that $|S| = n = s_r \cdot |A|$ and $S = \{s_1, s_2, ..., s_n\}$ such that $s_i \in A \, \forall \, i \in [1, n]$

The sampling ratio introduces a trade-off between the computational costs and the generator accuracy. If we use a $s_r$ too high, then the evaluation of weights at every edge addition step increases. If it is the other extreme, we risk using a small sample resulting in more randomness, which works against picking a preferential edge that a generator has defined using the weights.

The new edge $e_{ij}$ is then chosen stochastically using a probability $P_{ij}$ that is proportional to this weight $w_{ij}$ from the sample of edges $S$:

$$P_{ij} = \frac{w_{ij}}{\sum_{i',j' \in S} w_{i'j'}} \tag{1.1}$$

The value of $w_{ij}$ conferred to an edge is based on an arbitrarily complex mathematical expression and it is this expression that we represent using the *tree-based computer program*. Parent (non-terminal) vertices are operators while leaf (terminal) vertices are variables and constants. For the parent vertices, the vocabulary made available for constructing expressions include arithmetic operators $\{+, -, *, /\}$, general-purpose functions $\{x^y, e^x, log, abs, min, max\}$, conditional expressions $\{>, <, =, = 0\}$, and a special function $\psi$. Variables contain information specific to the two nodes (node $i$ and node $j$) involved in the edge ($e_{ij}$) considered: the sequential identifiers $\{i$ and $j\}$, in- and out-degrees of the two vertices $\{k(i,j), k_{in}(i,j)$ and $k_{out}(i,j)\}$, the undirected, directed and reverse distances $\{d(i,j), d_D(i,j),$ and $d_R(i,j)\}$ between the two vertices. In undirected networks, for the leaf nodes, the variables used are $k$, $d$, $i$ and $j$ apart from constants. The meaning of each of the variables is discussed below and summarised in table 1.2.

The *degree* centrality is the number of edges a node has. If the network is directed (meaning that the edges have direction and are not bidirectional), two different degree measures are defined, namely, in-degree and out-degree. In-degree is the number of edges directed towards the node, and out-degree is the number of edges directed away from the node. In such cases, the degree is the sum of indegree and out-degree. Let $A = \{a_{ij}\}$ be the adjacency matrix of a directed graph. If there exists an edge from node $i$ to node $j$ then $a_{ij} = 1$ else 0. The directed degree measures are given by:

$$k_{in} = \sum_m a_{mi} \qquad k_{out} = \sum_m a_{im}$$

The undirected distance measure $d(i,j)$ is the length of the shortest path consisting of edges between the two nodes considered. In the case of directed networks, the directed distance $d_D(i,j)$ between nodes $i$ and $j$ is the length of the shortest directed path from $i$ to $j$. The reverse distance $d_R(i,j)$ is the directed distance from nodes $j$ to $i$ i.e. $d_R(i,j) = d_D(j,i)$. These distances are calculated if such a path exists; if not, they are set to a high value (in this case, 11).

In a network of $N$ nodes, the sequential identifier for a node is its index $i \in \{1, ..., N\}$. The motivation for including this as a variable is to allow for the property that all nodes do not have to be (or behave) the same without regard to their topological properties. This non-topological property allows for *a priori* distinction in the types of nodes present in a network. This contributes a lot in explaining social networks [17]. For example, consider $w_{ij} = max(i, j)$; the probability of an edge forming between two nodes is higher for candidate edges when at least one node has a large identifier. [This intrinsic identifier might have been conferred based on the extent of outgoingness a person might have, thus leading to a proportional propensity to form friendships in a social network]. While this might seem simplistic, we can extend this index-based heterogeneity to make generators that can exhibit nodes with an affinity towards specific kinds of nodes, leading to introducing the affinity function $\psi$. It uses the modulo operation to partition the identifier space into a set of groups. It has three operands: a constant, g, which is the number of groups, and two conditional output expressions, a and b. If the nodes of the candidate edge i and j are equal modulo g, then they belong to the same group ("affinity"), and the function returns a. Otherwise, it returns b:

The *affinity* function $\psi$:

$$\psi_g(i, j, a, b) \equiv \psi_g(a, b) = \begin{cases} a, & if\ (i\%g) \equiv (j\%g) \\ b, & otherwise \end{cases} \quad (1.2)$$

For instance, consider a network of politicians in the United States Congress. The bipartisan system of Democrats and Republicans leads to two strong communities with a few edges between the two communities [20]. One of the generative mechanisms that could explain this network is $w_{ij} = \psi_2(10, 2)$. Essentially, this can be interpreted as a node with even/odd identifiers looking out more to form ties with other nodes with the same even/odd identifiers. This explains the homophilic tendency for democrats/conservatives to clump together more with people like themselves. In the last row, we also have a quantity called

| Variable | Name | Directed/Undirected |
|---|---|---|
| $i$, $j$ | Sequential identifiers | both |
| $k_i$, $k_j$ | Degree | both |
| $k_{in}$, $k_{out}$ | In- and Out-degrees | directed |
| $d$ | Shortest Distance | both |
| $d_D(i,j)$, $d_R(i,j)$ | Direct and Reverse Distance | directed |
| $\xi$ | Edge ratio | both |

Table 1.2: List of variable that can become part of the generator trees (the terminal nodes, specifically). There are also constants - integers from $\{0, 1, ..., 9\}$ and real numbers chosen randomly from $[0, 1]$

'edge ratio', this is an addition to the lexicon that we introduce which is not part of the original setup. It plays a very important part in describing time and we discuss its function later in section 2.3.1.

Hence, this minimal framework (Table 1.2 provides a vocabulary for specifying generators and expressions that represent and generate non-linear and non-centralized growth mechanisms.

## 1.2.2 The Fitness Function

Now given our target network and a generator, how can we answer whether the network created by the generator explains the network sufficiently well? Given the stochastic nature of the generation process, we must accommodate a gradient of deviation from the exact target network. We rely on simple yet fundamental aspects of the network's identity to measure similarity. We have to be careful here in our choice of the aspects because it is easy to smuggle in ad-hoc assumptions of the kind of generators we want explaining our networks through the back-door in the fitness function. We use the classical measures of degree centrality distribution of the network (in- and out-degree versions for directed networks), PageRank [21] centrality distribution (direct and reverse versions for directed networks), directed/undirected distances [22] and the triadic census [12].

The ubiquitous degree centrality distribution gives us information about which nodes are more/less connected. They are essential to studying a huge range of empirical and theoretical networks. From the Erdos-Renyi model, models using power laws and scale-free networks to world wide web, internet networks and social networks [23, 24, 25].

The PageRank centralities give us information about the influence or importance of nodes whose reach extends beyond their immediate neighbours. Let $A = \{a_{ij}\}$ be the adjacency matrix of a directed graph. The direct PageRank centrality $x_D(i,j)$ of node $i$ is given by:

$$x_D(i,j) = \alpha \sum_m \frac{a_{mi}}{k_{out}(m)} x_D(m,j) + \beta$$

where $\alpha$ and $\beta$ are constants and $k_{out}(m)$ is the out-degree of node $m$ if such degree is positive, or $k_{out}(m) = 1$ if the out-degree of $m$ is null. Reverse Pagerank centrality is gotten analogously by reversing the direction of the edges and recalculating.

The directed/undirected distance between two vertices in a graph is a simple but surprisingly useful notion. It has led to the definition of several graph parameters such as the diameter, the radius, the average distance and gives a quasi-metric instead of relying on an extrinsic coordinate system or embedding in some dimensional space [26].

The use of triadic census originates from the observation that there is a meaningful division of networks, especially social and communication networks, into smaller classes, where the class of the network is determined by its generating process.Each class is distinguished by a common structural element: a temporal-topological network motif that corresponds to the network representation of communication events in that network class. Only specific network topologies are conceivable inside a dynamic class because of these underlying features. In this way, the possible three node structures (triads) lend themselves to become an integral part of nailing the identity of a network through its generative process [27, 28]. In our work we modified the algorithm proposed in Ref. [29], by reducing the need for overcounting thus resulting in a lower running time albeit with the same time complexity - $\mathcal{O}(N^2)$.

These attributes which are a mix of simple as well as more detailed aspects of the structure are now reduced to metrics by calculating the dissimilarities between the respective distributions. The degree and Pagerank centrality distributions employ the *Earth Mover's Distance* [30]. As for the more complicated distance and triadic pattern distributions, a simpler ratio-based dissimilarity metric is calculated. The degree and PageRank centrality distributions have a well defined notion of distance between the bins. However, for something like the triadic census where we are comparing the difference in the 16 kinds of triads between two networks, a simpler ratio-based dissimilarity metric is calculated so that we can normalise and add up the differences in different quantities and obtain one value for each attribute (More in Appendix A.2).

Reducing the dissimilarities of all these metrics would then result in the synthetic and input network to become more similar. Reducing the dissimilarity for one of the metrics might result in an increase in another *i.e.* we have a multi-objective optimization problem. The way this is handled is to first enable direct comparison of the different dissimilarities by normalising them and then decide on how we shall go about minimising the 'overall' dissimilarity. The normalisation is done by looking at the dissimilarities with respect to their improvement in explanatory power compared to a null model. We divide the dissimilarity between the generated and the target networks by the dissimilarity between the generated network and the average of 30 Erdos-Renyi random networks (with the same number of nodes and edges as the target network). We now have a scale from 0 to 1 correspondingly going from "generated network exactly equivalent to the target network" to "generated network explaining the target network no better than an E-R random network". In `synth` specifically, the algorithm gets at improving generator quality by minimizing the highest of these ratios. It is important to note that the fitness values are not comparable across different networks and cannot be used as an absolute measure of quality.

## 1.3 An Illustrative Example

The evolutionary algorithm begins with the definition of the population of solutions. The choice of solutions and the quantities they optimise are of utmost important. The choice of one solution is straightforward - simply optimise for the solution with the lowest fitness value. Ref. [31] discovered that fitness-based selection increases the propensity for solutions to expand in size. They argue that utilising a fixed fitness function with a discrete but variable length representation leads to such expansion. The goal of simple static evaluation searches then is to locate trial solutions that are as fit as existing trial solutions. In general, variable length allows for many more long than short representations of a particular solution. As a result, we expect longer representations to appear more frequently in search (without a length bias), and hence representation length to tends to rise. Bloat is the result of fitness-based selection and to control this we use the anti-bloat tolerance ratio.

There are two solutions maintained: $w_b$ - one that contains the generator with the best fitness (*i.e.* the one with the fitness closest to 0); and the other solution $w_s$ contains the generator that has the shortest tree size with fitness that is at most $b_r\%$ worse from the best solution. This parameter $b_r$ is the "*anti-bloat tolerance*". This length bias exists to ward off bloat or redundant structures within the generator tree (refer Appendix A.1). This anti-bloat ratio pressures the evolutionary process into producing generators that are small and closest to their Kolmogorov complexity [32]. Here, the generator size neatly provides for an upper bound in the Kolmogorov complexity.

Let us begin by looking at the creation of a network for a given network generator and then zoom out to look at the entire evolutionary algorithm describing the search for a network generation process for a given input (or target) network. Follow along using Fig. 1.1. For the sake of example, as the input network to `synth`, let's use the directed linking patterns and discussion topics network of political bloggers (conservatives and liberals) in the 2004 US elections [33].

We initialise a random generator (Appendix A.3) $w_r$ (happens to be $= e^{(4-k)}$, say) and set $w_r = w_b = w_s$. For every evolutionary search step we randomly pick one of the solutions from our population $w_b, w_s$ and clone it $w_c$. We now mutate the generator tree of the clone *i.e.* we replace a sub-tree of $w_c$ with another randomly created generator tree. Say, the sub-tree $k$ was replaced with the randomly generated tree $2 * d$. We create the synthetic network using the generator formula given in $w_c = e^{(4-2d)}$ (Fig. 1.1, Fig. 2.4). We calculate the dissimilarity between this network and the input political blog network, if the fitness is better *i.e.* if its value is closer to 0 as compared to the networks generated using $w_b$ and/or $w_s$, then $w_c$ replaces one or both as the new best fitting and/or shortest generators in the population. This is one evolutionary search step. We repeat these steps for a long time, more specifically, we keep searching until the population of solutions has not improved in the past 1000 search steps.

In this example of the directed political blogs network, over many runs the consistent generator obtained was actually $w_b = e^{(4-2d)}$. This can be interpreted by looking at the fact that the weight given to an edge exponentially decays close (but never equal) to zero even for undirected distances of 3 and greater. Since our candidate edge samples are all (ordered) node pairs which do not have an edge yet, a high weight to $d = 1$ implies that the generator prefers to add an edge ($e_{ij}$) to a node pair which has the reciprocal edge $e_{ji}$, that is reciprocity is highly preferred (tendency to connect to cite/refer/link to another blog that did the same to your blog). This distribution is enough to reconstruct the two distinct (liberal and conservative communities) that were observed in the empirical network. Edges between blogs with a low undirected distance are more likely, but links to blogs further away are also viable (albeit lesser and lesser). As a result, the generator is able to reconstruct the two distinct communities observed.

## 1.4  Next Steps

Now that we have set up the playing field, it is instructive to note where `synth` falls short and how we will proceed to work with the limitations in line with our goals mentioned in section 0.2.

**Limit 1** The nature of the algorithm is that it doesn't account for network growth. For instance, it does not account for node additions/deletions, all of the nodes have to be initialised.

**Limit 2** The algorithm is defined to generate probabilistic models so a static fitting model like the stochastic block model cannot be exactly fit. A lattice like network with its high orderedness might not be easily captured.

**Limit 3** While the framework was proposed as a proof of concept with a simple lexicon. It does not have the language to describe more special kinds of networks - notably, time-varying networks. This is a huge class of networks and a major part of the next chapter is devoted to overcome this hurdle.

**Limit 4** Even if we try to introduce more specificity to the context by introducing relevant variables and operators, we will end up with a large search space. This will lead to an increased computational effort to retrieve solutions. This is dealt in sections 2.6, 3.1. Apart from an increase in search time, the algorithm which was initially implemented in Java was ported to Python for more accessibility. It now needs to become faster to be usable for networks of large sizes (Appendix A.4).

Evidently, `synth` does not cover all of the double dichotomy table. A recent paper [34] found that the most efficient models that retrieve structure are stacked machine learning models. So it is important to note that there always *are* going to be certain kinds of networks that are not going to work well with `synth` despite our best efforts to extend the algorithm.

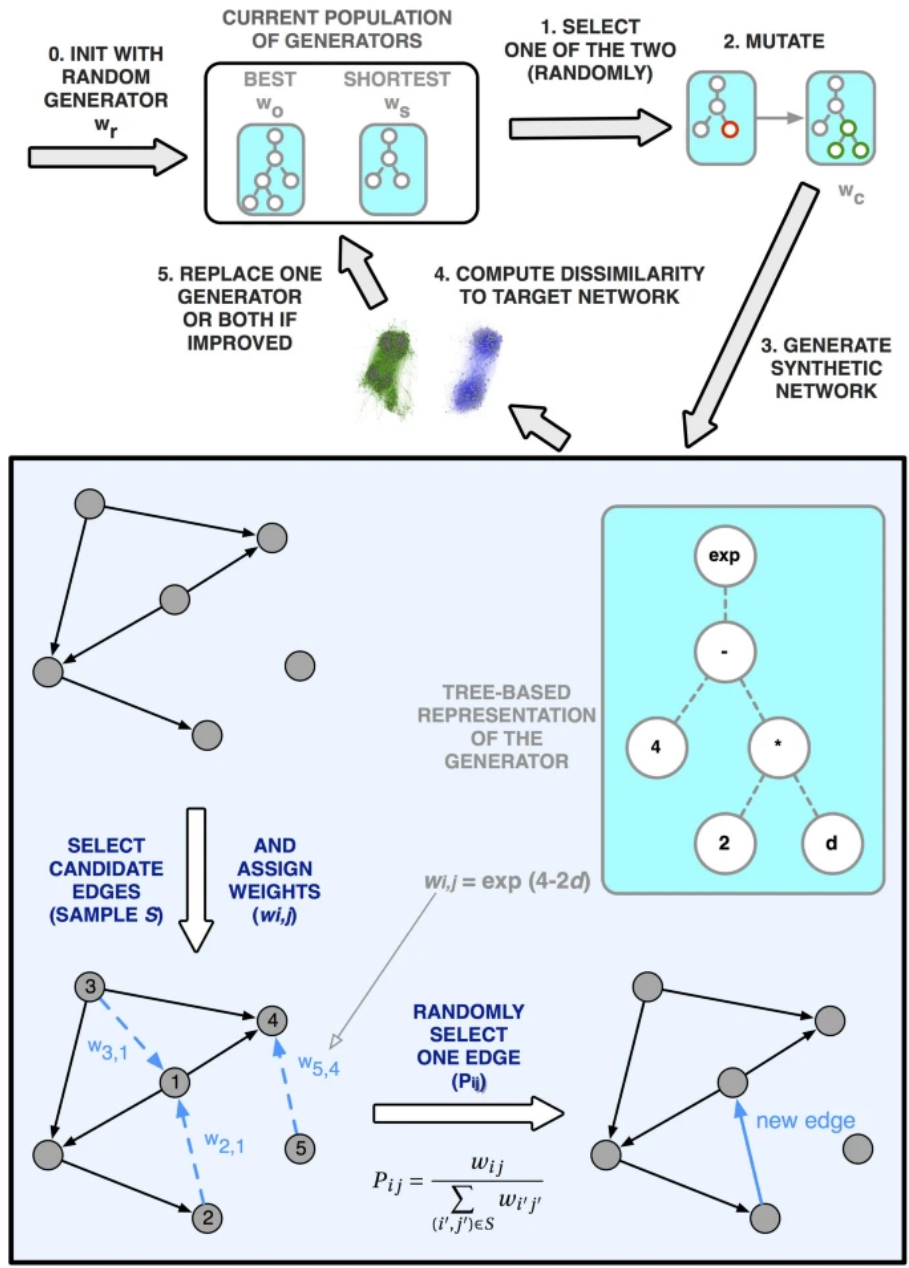

Figure 1.1: Automatic discovery of models: "Evolutionary loop including the synthetic network generation process. The top part of this figure describes evolution at the generator population level, while the bottom (framed) part describes the evolution of a network for a given generator." - Figure 1 from Ref. [9].

# Chapter 2

# Dynamic Networks & `synth`

## 2.1  A naive extension of "Dynamic Networks"

As put forward in section 0.2 and motivated by Limit 3, one of our objectives is to extend the framework to dynamic networks. There are different ways evolving networks have been analysed, such methods are neatly summarised in Ref. [35].The simplest way to introduce an element of a network changing is observing at multiple stages of development. Therefore, we could introduce multiple snapshots - taken at different points of the growth of the network and present it to `synth` and see if we can retrieve the growth process in the form of a generator expression that explains/fits all of the snapshots. We started with 5 snapshots. Intuitively, we might expect that more information would mean it would be easier for `synth` to retrieve a rule but we obtained solutions with fitness near 1, which means that they were only just as good as the Erdos-Renyi random network in explaining the target network.

The reason is that we are expecting to find a solution that is better than not only one fitness function but simultaneously five. Given the large and independent search spaces for each of the snapshots, it is highly likely that the search get stuck at a local optimum because of the strict conditions. Currently, the algorithm has no information that the snapshots are causally connected. We can significantly reduce redundancies in the search spaces, if we introduce information like the time stamp of the snapshots, the aspects of the generator that does not change between the many snapshots. Nevertheless, this prompted us to try relaxing the mutation threshold, we could see it finding more rules but they were still not better than random [Fig. 2.1]. We decided to restrict the number of snapshots to two - one at the halfway mark and one at the end, thereby giving us a relatively smaller search space to work in. This change did not result in solutions better than random either. It is very much possible to go in the direction of causally connecting snapshots and making an overall fitness function that uses all of the snapshots together like in Ref. [36] but we will be choosing to complicate the framework (like introducing external parameters) without sufficient preliminary evidence. Which is against the symbolic regression idea of relying as much as possible on local information. Therefore, while we retain the idea of multiple snapshots wherever relevant we will continue looking for other methods to restrict the search space or find another way to define and analyse dynamic networks.

## 2.2   Redefining the network generative process

In an effort to try restricting the search space as well as overcome one of the shortcomings of the algorithm - we decided to change the way in which nodes were added into the network. Initialising and defining all the nodes at the start creates $\binom{N}{2}$ possible edges to pick right from the beginning. We wanted to check if this might be leading to large initial stochasticity. We wanted to know if this randomness was avoided whether the multi-snapshot version of `synth` will now be able to recover solutions. So instead of initialising all nodes at the beginning, we started with $m_0$ initial nodes and at every time step, one new node was initialised and $m \leq m_0$ edges were added using probabilities from the same generator expression in Eq. 1.1. Note that this also tries to address Limit 1 by accommodating a dynamic node addition process akin to the edge addition we've been carrying out. It was observed that the same generator rules when used in this formulation of the algorithm lead to radically different networks which didn't resemble each other.

A simple way to contrast the effects of the two kinds of generators *i.e.* the one with and without dynamically adding nodes into the network, is to compare the properties of resultant networks created by using the same generator equation $w_{ij}$. We looked at the two most common generators - the Erdos-Renyi ($w_{ij} = c$) and Preferential Attachment generators ($w_{ij} = k$) and compared their degree distributions in Fig. 2.2. We see that the two methods of node initialisation (static and dynamic) give out drastically different degree distributions. More interesting is that the PA and ER networks created using the dynamic method resulted in similar distributions even though they have distinct propensities to attach to nodes. We observed that the dynamic method when used on different generators resulted in networks that had similar degree distributions (and dissimilarities). One explanation for this similarity caused by the dynamic method is that it had restricted the space of possible networks too much. Because of the small set of possible edges to choose from in the beginning, no matter the generator rules used the probability distribution was roughly the same, this explains the similar degree distributions of the ER and PA generators (and other dissimilarity values) when using this transmogrified dynamic version of the algorithm.

We cannot proceed with this network growth model in the context of symbolic regression because it fundamentally alters the way in which we base the identity of our network, resulting in an apples to oranges comparison. The meaning of the preferential attachment or the random network models or any generator we choose to use becomes incongruous to the static model. Not only that, since the distributions are very similar, they are not useful as generative models providing distinct signatures either. It also detracts from the idea of a unifying generative process put forward in Ref. [9]. We can see how the network growth model cannot be straightforwardly incorporated with the existing framework without a major overhaul of the dissimilarity metrics, the fitness functions and/or the vocabulary of the generators used. While this strategy did not pan out well, trying to define what we meant by dynamic networks lead to a formulation of dynamic networks that was right under our nose

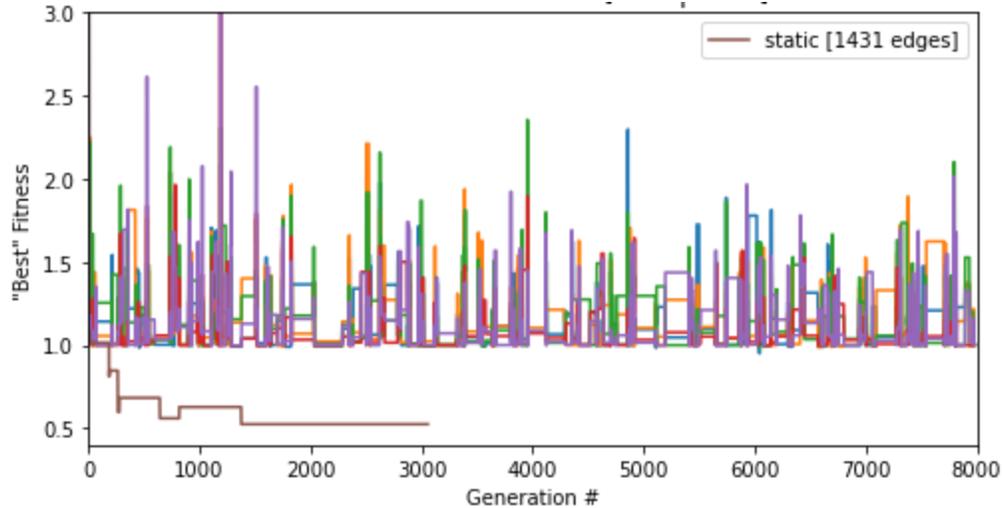

Figure 2.1: A 200N‖1431E network created using the $PA$ generator was run through `synth` to test its abilities to recover the generator but with the information of 5 network snapshots (instead of the usual 1 snapshot) at different stages of creation (720, 850, 960, 1200, 1431 edges). The best solution of the search is updated if the worst of the 5 solutions at each snapshot is improved. The fitness of the best solution for each of the 5 snapshots has been averaged over 10 runs and plotted against the number of evolutionary search steps (generation #). Plotted for comparison is the single snapshot `synth`'s recovery of the same network averaged over 10 runs and it arrives at the correct generator while the multi-snapshot version struggles to find even one solution that is better than the ER generator (fitness < 1)

all this time.

## 2.3 An intrinsic measure of time

### 2.3.1 Edge Ratio

In order to avoid confusion, we make it explicit that we went back to the original version of the algorithm where we initialised all the nodes of the network at the start.

To peer under the dynamic structure of networks, an index of time needs to be defined. There are many ways in which dynamic networks can be defined and represented. They could be defined using discrete/continuous time and represented as an aggregated, a time-ordered or time-varying graphs [35]. Using a timeline that indicates the edges present in a network at a given time results in additional extrinsic temporal information that can very majorly bias our model depending on the timescale [37]. From our previous encounter with the node addition model and in the spirit of the symbolic regression approach's reliance on only local information and parameter-free fitness functions, we decide to use the number of edges currently in the network normalised with respect to the number of edges present in the final observation of the network as an index of time [36]. The generative processes we are already dealing with anyway add edges one step at a time, so a natural proxy for the time variable is to now use the number of edges currently added to the network. We define the edge ratio $\xi$ as the number of edges currently in the network (e) by the total number of edges (E) we want in the network at the end after all edge additions ($\xi = \frac{e}{E}$). Note that this 'edge ratio' $\xi \in [0, 1]$ and because we choose to focus only on dynamic networks that don't have any edge deletions, the edge ratio is a non-decreasing quantity. Suppose we have to create a network of 100N‖1000E (100 nodes and 1000 edges) using some generator, $\xi = 0$ is the stage when all 100 nodes are initialised without any edges added, $\xi = 0.5$ is when 500 edges are added to the network and $\xi = 1$ is when all 1000 edges are added.

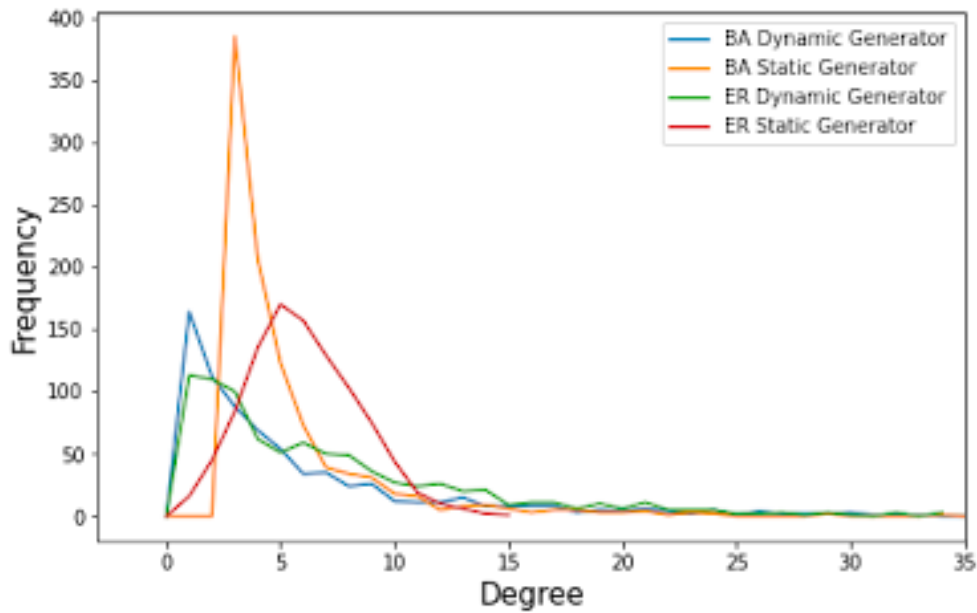

Figure 2.2: Comparison of the degree distribution of the 1000N|2991E networks generated using the PA/ER generators in the original `synth` algorithm (labelled as PA/ER static) and the version with nodes initialised one step at a time (labelled as PA/ER dynamic). We see very similar distributions of the ER dynamic and PA dynamic curves while the ER and PA static are their usual distinct distributions and completely different from their counterparts.

Now that we have enabled $\xi$ to become part of the vocabulary, we have the machinery to create a host of time-dependent rules and analyse dynamic networks.

### 2.3.2 Delta Function

Novel generators or existing ones with weights dependent on the edge ratio can now be created. However, we were interested in probing into a kind of problem that is strangely familiar and at the same time is woefully understudied in the literature.

With generator expressions representing networks, we wonder how will networks created using not one, but multiple generators look like? Suppose a network is created using one generator up till a point and then we switch over to another generator for the rest of the creation process, can such forms of multi-stage network generation process be somehow captured? We introduce into our generator vocabulary a function that can capture this phenomenon. This function $\Delta$ indicates that until a particular edge-ratio $\xi$, generator $a$ is used to add new edges after which, generator $b$ is used. ¡insert citation of paper on networks with multiple processes¿.

The $\Delta$ function is given as:

$$\Delta_g(Gen_1, Gen_2) = \begin{cases} Gen_1, & if\ g \leq \xi \\ Gen_2, & otherwise \end{cases} \quad (2.1)$$

A generator of the form $w_{ij} = \Delta_{0.5}(c, k(i,j)) \equiv \Delta_{0.5}(ER, PA)$ means that we use the E-R random generator until half of the target number of edges are added and the rest of the edges are added using the degree PA model.

We test whether the $\Delta$ generator gives out networks that are consistent and similar to each other, so as to warrant their own distinct generator, because there is no guarantee that using two different generators which have their own distinct properties will lead to a final network that also has a distinct signature. For all we know, the $\Delta$ function with two specific generators could end up looking like an E-R random network through our dissimilarity measures. While that may necessitate changes in our fitness function it also gives the framework an opportunity to bring in more kinds of processes that can be reconstructed by just using the structure of the network.

To check the similarity of the $\Delta$ function, we generated 5 400N‖4000E networks using $w_{ij} = \Delta_{0.5}(PA, PA')$ and each of them were compared with 5 sets of networks generated using $w_{ij} = \Delta_{0.5}(PA, PA')$. At the end of the network generation process ($\xi = 1.0$), we obtain a low fitness score of 0.178 (table 2.1) indicating that the generator is distinctive while still giving rise to different networks due to the inherent but enhanced stochasticity due to two different generators on top of each other. If one notices, the $\Delta_x(Gen_1, Gen_2)$ is a shorthand and this action can be replicated using the generator $w_{ij} = (> \xi\, x\, Gen_2\, Gen_1)$. This is read as "If $\xi > 0.5$ use $Gen_2$, else use $Gen_1$. We introduce this shorthand function into the vocabulary so that it is easier to access this kind of a generator and in a way we are assigning equal importance to this function as that of any other function. We wanted to test whether this function is performing as expected and as a consistency check we repeated our comparison of the 5 networks with 5 sets of networks generated using $w_{ij} = (> \xi\, 0.5\, PA'\, PA)$. The comparison with each set is done at $\xi = 0.5$ and $\xi = 1.0$ and their corresponding mean fitness and dissimilarity values are tabulated in table 2.1. We see that this second set of comparisons also has a similar mean fitness values of 0.159. Furthermore, we also make a third comparison with 5 sets of networks generated using $w_{ij} = PA$. This was to see whether the fitnesses at $\xi = 0.5$ are comparable, which is until when all three sets of comparison are identical. At $\xi = 1.0$, we expect the first two comparison to yield fitness and dissimilarity values better than comparing with just $PA$ generator. We see similar fitness values for all three comparisons at $\xi = 0.5$ - 0.511, 0.579 and 0.568. At $\xi = 1.0$ we see fitness values of 0.178, 0.159 and 1.28, the large value of 1.28 confirms that the $\Delta$ function, at least for this combination of generators, has a distinctive signature that is not its constituent generators. In the table, the results for these tests repeated with 300N∥3000E networks are also reported and we can infer the same irrespective of their different sizes.

## 2.4 Effect of initial conditions

Apart from inspiring this multi-generator dynamic network, the node initialisation detour also made us think about the role of initial conditions and whether they affect the network generated on top of a given initial network. To test this, we created the first part of a network with either the PA or the ER generator, and on top of that we created the remaining part of the network using another generator. `synth` was modified such that it is given the target network and it starts off with this initial network with the goal of finding a generator for the remaining edges *i.e.* to see if the second part could be reliably retrieved irrespective of our initial conditions. We are able to exactly retrieve augmenting generators belonging to the families - ER/PA/PA'/ID [1]. This implies that a network generated using this two-part process can be extracted generator by generator provided we know the point at which the switch happens.

This does not imply that initial conditions do not matter at all. For instance, it is not a seemingly reasonable expectation to be able to use and retrieve an augmenting generator like the SC-$\delta$ family [17] which uses $w_{ij} = \psi_2(9^i, 9^9)$ and has a distinctive two-community structure. In fact, we used the Girvan-Newman algorithm [38] and found no non-trivial two community structure of the network with SC-$\delta$ as the augmenting generator and PA/ER as the initial generator. So this means that the initial condition can sometimes destroy the characteristic of the augmenting generator. (We have more to say about this in section 2.6.) However, this does not rule out the possibility that `synth` finds a different solution that explains this final network well. This is a cautionary step against entirely relying on solutions that follow the principle of Occam's razor (we reduce this reliance in Appendix A.1).

It also brings to the foreground the possibility that two distinct generators giving rise to dissimilar networks when used as augmenting generators to some initial generator could give rise to very similar networks, because of the stochasticity added to the network structure by the initial generator. Unless the two phased network we create is distinctive **because** of the two generators in use then we are always bound to the possibility of getting generators that are not of such a two-part form that might still be able to explain our target network and maybe even do it better. This is a counter step to our previous paragraph justifying the need for why Occam's razor still has a place in our framework (not to mention the increased computational costs if we do away with preferring shorter solutions).

---

[1] The weight distribution $w_{ij}$ of - ER: constant, PA: degree of origin/target node ($k_i$, $k_j$), PA': $k^k$, ID: the sequential identifier ($i$ or $j$)

## 2.5 The search begins

Now that we know that `synth` can retrieve the augmenting generator given the initial generator, we proceeded to shed one of the crutches we gave to it. We no longer provide the value of $x$ in $\Delta_x(Gen_1, Gen_2)$. So the network generation begins right from edge #0 instead of generating edges only for the second part of the process. However, we chose to provide `synth` two snapshots as a way of giving it more information and we choose to update the current best/best-fitting solution if the fitness function is improved for both the snapshots.

We started with the target network generated by $\Delta_{0.5}(PA, ER)$. All the runs yielded the same solution - PA. This explains the first part exactly, and the second part too in some sense. Because, the dissimilarity between ER networks and the ER generator are always nearly 1 (due to the way we normalise our dissimilarity metrics using ER networks). Given `synth`'s built-in preference for simpler solutions this is more likely to be found than the $\Delta$ generator. Now we need to be sure whether `synth` can potentially identify the two constituent generators and moreover if it can even find the point where we switch from one to another. In order to test that we use a $\Delta_{0.5}(Gen_1, Gen_2)$ generator which has low values (closer to zero implies more identical to target network) of fitness at both of the two snapshots at $\xi = 0.5 \,\&\, 1.0$ indicating a distinctive character that isn't affected by stochasticity. We found PA and the PA' generators, two of the more commonly found families, to qualify as both $Gen_1$ and $Gen_2$. Even after making these changes, for target network generated using $\Delta_{0.5}(PA, PA')$, all of the solutions `synth` retrieved happened to belong to the PA family. The first snapshot was fit well while the second snapshot (at $\xi = 1.0$) performed only marginally better than random.

We initially worried if it might be due to finite size effects, but networks of size 100N‖1000E to 500N‖5000E were all tried, and no difference was found in the solutions and the corresponding fitnesses with which the obtained solutions explained the target networks. We made sure to check that the choice of switch-point did not affect the distinctiveness of the two generators used thus not compromising its genotypic signature that `synth` can try to detect

[Fig. 2.3].

We decided to then reduce the tolerance from 0.10 to 0.05 per snapshot to allow for more complicated solutions albeit with better explaining power. Note that tolerance is the amount of reduction in fitness tolerated by each snapshot in favour of a simpler solution. Even after this change, all of the solutions found by `synth` turned out to be the PA generator, **all except one**. This exception happened to be the exact form of the solution but with the switch-point to be a bit off: $\Delta_{0.645}(Gen_1, Gen_2)$. This reassured us that it was theoretically possible to get back the generator of the input network.

## 2.6 A little help from Recombination

While we were able to recover the $\Delta$ generator once, we need to be able to retrieve this solution more consistently, and for that a more efficient foraging of the search space was needed. While the extension of the vocabulary has enriched the possible kinds of generators for creating a network it has further increased the multi-dimensional search space of the generators making it harder for the algorithm to efficiently find the generator that explains any given network. With only two generators in the population we find that one dominates the other due to our preference for simpler solutions and therefore we need a mechanism that will mix up the solutions leading to varied parts of the search space being accessed. Introducing recombination of solutions from a larger population of generators is straight from the playbook of genetic programming which we have seen – although not in our context – work out in instances before [39].

We tested the effect of recombination on `synth`'s ability to recover solutions in the single snapshot iteration of the algorithm. So far, the population of generators consisted of only the best and the best-fitting solution and we had reasoned why in section 1.3. From our runs, we find that the evolutionary search process can get stuck for a long time without improvement on one solution. It is usually due to one among the quantities in our fitness function that we calculate the dissimilarity for(section 1.2.2). While the rest of the quantities have a lower dissimilarity, the one quantity that has a relatively larger value hinders the search process. All one has to do is to compare the differences between the mean fitnesses and the mean dissimilarities in table 2.1 (0.511 vs 0.263, 0.178 vs 0.060, etc). In order to offset this hindrance to enable `synth` to continue finding solutions, a third generator was added that optimised for the lowest mean dissimilarity of the generated network compared to the target network [2].

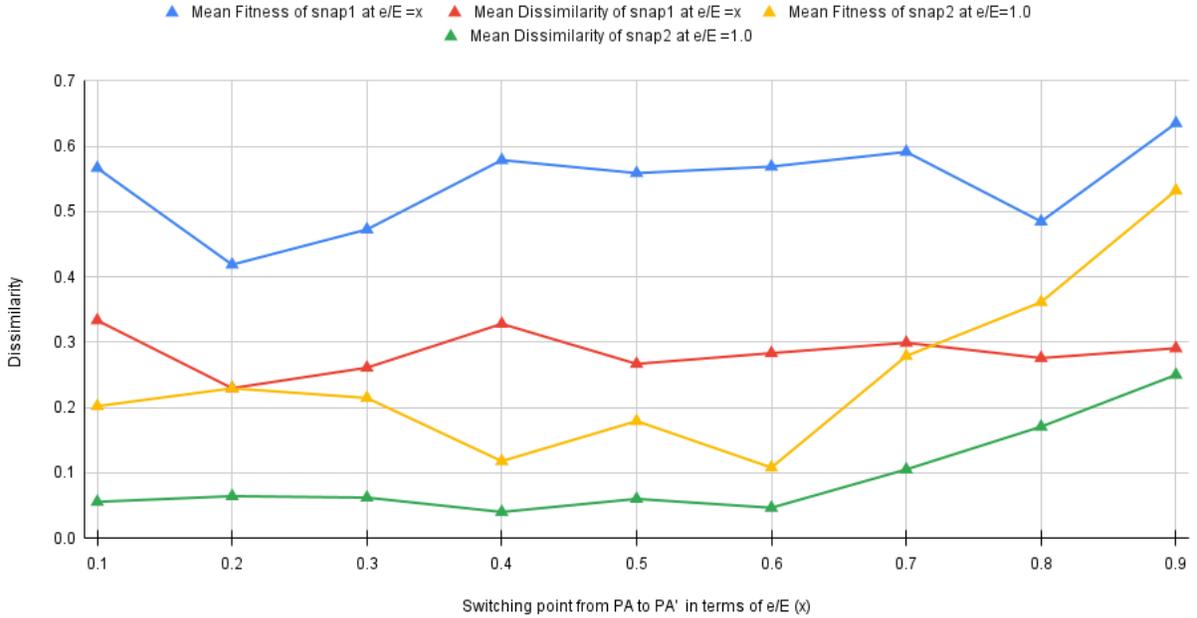

Figure 2.3: Mean dissimilarity and fitness values are reported for snapshots 1 taken at $\xi = x$ [where x goes from 0.1 to 0.9 in intervals of 0.1] and for snapshot 2 taken at $\xi = 1.0$. For a given y-axis value, the concerned quantity is averaged over 5 300N‖3000E networks with snapshot 1 taken at $\xi = x$ (the x-axis value) and snapshot 2 taken at $\xi = 1.0$. The network generator used is $\Delta_x(PA, PA')$. For a given x-axis value, we have plotted the mean fitness/dissimilarity for both snapshot 1 and snapshot 2. We see that switching from one generator to the second for $\xi < 0.7$ will maintain both the generators' distinctive contribution to the network structure and not devolve into a ER network because their corresponding fitness/dissimilarity values are fairly stable and low.

We found that on an average adding recombination lead to the framework finding the exact solutions quicker than without recombination at least for input generators belonging to the families ER, PA, PA', $d$ [3]. The results could be made stronger by increasing the generator population, which would involve finding meaningful quantities to optimise which would result in finding better fitting solutions [spoiler alert: 3.1]. However, this will come at increased computational costs.

Now we turn to the multi-snapshot version of `synth` and we extend the choice of generators available to recombine by introducing two new solutions - one that optimises for the best fitness of snapshot 1 and one for snapshot 2. We randomly choose to either mutate or recombine, and if we choose the latter we randomly pick two of the four generators from the population. We choose an arbitrary sub-tree in the first generator and replace it with a similarly chosen sub-tree from the second generator. For the input generator $\Delta_{0.5}(PA', PA)$, `synth` was able to retrieve the constituent generators as well as get close to the switch-point in the majority of the runs. Table 2.2 displays the results of 5 runs of generator retrieval using the $\Delta_{0.5}(PA', PA)$ generator using the two-snapshot version of `synth`. Since we ran this recovery for only the $PA'$ to $PA$ generator, we proceeded to run it for multiple pairs of generators. We have used representative generators from different families as detailed in Ref. [17]. We have tabulated whether a certain pair of generators using the $\Delta_{0.5}(Gen_1, Gen_2)$ are recovered more than once, only once or not recovered in table 2.3. We see that nearly half of the pairs tested are recovered. We see that $SC - \beta$ ($w_{ij} = \psi_3(2^k, 0.5)$) and $SC - \alpha$ ($w_{ij} = \psi_7(k^3, 4)$) generators are not recoverable that easily. Most curiously, we see a lot of $\Delta$ generators with $SC - \delta$ ($w_{ij} = \psi_2(9^i, 9^9)$) as the initial generator being retrieved but when used as the augmenting generator.

---

[2] A further observation was that it was the undirected distance quantity in the fitness function that caused the process to get stuck. Therefore, we introduced a fourth solution in the population that optimised for solutions with the lowest dissimilarity in undirected distance

[3] The weight distribution $w_{ij}$ of - ER: constant, PA: degree of origin/target node ($k_i/k_j$), PA': $k^k$, $d$: undirected distance between origin and target nodes.
39

| Network size 400N‖4000E Generated using $\Delta_{0.5}(PA, PA')$ | | Comparing Fitness of Input Network and networks generated with: | Input Network #1 | Input Network #2 | Input Network #3 | Input Network #4 | Input Network #5 | Average Quantity over 5 input networks (x 10 runs each) |
|---|---|---|---|---|---|---|---|---|
| $\xi = 0.5$ | Mean Fitness (over 10 runs) | $\Delta_{0.5}(PA, PA')$ | 0.467 | 0.465 | 0.518 | 0.511 | 0.595 | 0.511 |
| | | $(> \xi\, 0.5\ PA'\ PA)$ | 0.499 | 0.793 | 0.462 | 0.584 | 0.557 | 0.579 |
| | | $PA$ | 0.529 | 0.57 | 0.493 | 0.694 | 0.551 | 0.568 |
| | Mean Dissimilarity (over 10 runs) | $\Delta_{0.5}(PA, PA')$ | 0.222 | 0.223 | 0.259 | 0.297 | 0.313 | 0.263 |
| | | $(> \xi\, 0.5\ PA'\ PA)$ | 0.244 | 0.297 | 0.252 | 0.311 | 0.302 | 0.281 |
| | | $PA$ | 0.251 | 0.236 | 0.264 | 0.313 | 0.269 | 0.267 |
| $\xi = 1.0$ | Mean Fitness (over 10 runs) | $\Delta_{0.5}(PA, PA')$ | 0.087 | 0.248 | 0.232 | 0.151 | 0.171 | 0.178 |
| | | $(> \xi\, 0.5\ PA'\ PA)$ | 0.048 | 0.226 | 0.236 | 0.215 | 0.072 | 0.159 |
| | | $PA$ | 1.28 | 1.36 | 1.239 | 1.236 | 1.283 | 1.28 |
| | Mean Dissimilarity (over 10 runs) | $\Delta_{0.5}(PA, PA')$ | 0.033 | 0.076 | 0.073 | 0.06 | 0.056 | 0.06 |
| | | $(> \xi\, 0.5\ PA'\ PA)$ | 0.026 | 0.069 | 0.074 | 0.075 | 0.031 | 0.055 |
| | | $PA$ | 0.968 | 0.999 | 0.966 | 0.968 | 0.98 | 0.976 |

| Network size 300N‖3000E Generated using $\Delta_{0.5}(PA, PA')$ | | Network compared with Input Network: | Input Network #1 | Input Network #2 | Input Network #3 | Input Network #4 | Input Network #5 | Average Quantity over all 5 input networks (x 30 runs each) |
|---|---|---|---|---|---|---|---|---|
| $\xi = 0.5$ | Mean Fitness (over 30 runs) | $\Delta_{0.5}(PA, PA')$ | 0.461 | 0.489 | 0.574 | 0.574 | 0.561 | 0.532 |
| | | $PA$ | 0.479 | 0.478 | 0.541 | 0.625 | 0.501 | 0.525 |
| | Mean Dissimilarity (over 30 runs) | $\Delta_{0.5}(PA, PA')$ | 0.289 | 0.263 | 0.286 | 0.327 | 0.312 | 0.295 |
| | | $PA$ | 0.306 | 0.262 | 0.272 | 0.327 | 0.293 | 0.292 |
| $\xi = 1.0$ | Mean Fitness (over 30 runs) | $\Delta_{0.5}(PA, PA')$ | 0.238 | 0.238 | 0.101 | 0.09 | 0.143 | 0.162 |
| | | $PA$ | 1.587 | 1.768 | 1.696 | 1.773 | 1.772 | 1.719 |
| | Mean Dissimilarity (over 30 runs) | $\Delta_{0.5}(PA, PA')$ | 0.075 | 0.074 | 0.04 | 0.041 | 0.057 | 0.057 |
| | | $PA$ | 1.038 | 1.101 | 1.073 | 1.095 | 1.097 | 1.081 |

Table 2.1: 5 400N‖4000E networks were generated using $w_{ij} = \Delta_{0.5}(PA, PA')$ and each of them were compared with 5 sets of networks generated using - 1) $w_{ij} = \Delta_{0.5}(PA, PA')$, 2) $w_{ij} = (> \xi\, 0.5\ PA'\ PA)$, 3) $w_{ij} = PA$. The comparison with each set is done at $\xi = 0.5$ and $\xi = 1.0$ and their corresponding mean fitness and dissimilarity values are tabulated. The tests were repeated with 300N‖3000E networks as well.

| Run | Best Fitness at $\frac{e}{E} = 0.5$ | Best Fitness at $\frac{e}{E} = 1.0$ | Best Generator (Pruned) |
|---|---|---|---|
| 1 | 0.1755651518 | 0.1789836148 | $k_j - (1.7 * k_i) - 2.1$ |
| 2 | 0.06416444098 | 0.04665495647 | $k_i * \Delta 0.46(PA', 6)$ |
| 3 | 0.04477379303 | 0.06425441079 | $\Delta_{0.55}(PA', PA)$ |
| 4 | 0.07882695179 | 0.02365060356 | $\Delta_{0.45}(PA', PA)$ |
| 5 | 0.04975868453 | 0.04799432543 | $\Delta_{0.47}(PA', PA)$ |

Table 2.2: Symbolic Regression on 300N‖3000E input network generated using $\Delta_{0.50}(PA', PA)$. We see that majority of the runs retrieve the $\Delta$ generator and have gotten the switch point close to the correct value of $\xi = 0.5$ as well.

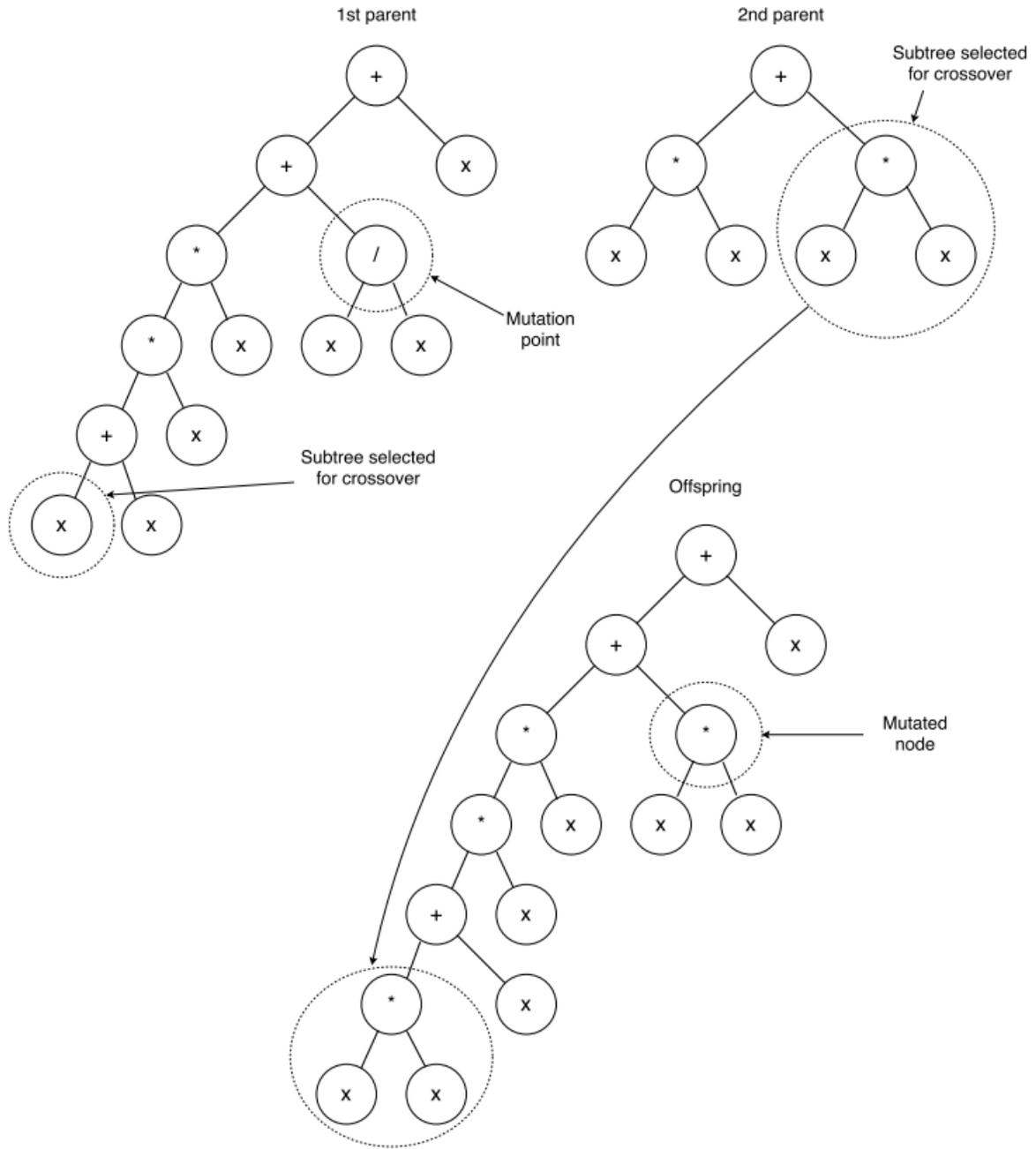

Figure 2.4: Illustration of the (headless chicken) mutation and recombination processes. Parent 1 generator: $w_{ij} = ((((x+x)*x)*x)+(x/x))+x$. Parent 2 generator: $w_{ij} = (x*x)+(x*x)$. Offspring generator: $w_{ij} = (((((x*x)+x)*x)*x)+(x*x))+x$. Figure 3.3 from Ref. [40]

| $Gen_1$ | $Gen_2$ | Recovered exact solution? | Best Fitness | Best Mean Dissimilarity |
|---|---|---|---|---|
| **ER** | **PA** | **Yes** | 0.43 | 0.35 |
| **ID** | **ID'** | **Yes** | 0.13 | 0.05 |
| **ID'** | **ID** | **Yes** | 0.08 | 0.03 |
| **ID'** | **PA** | **Yes** | 0.11 | 0.04 |
| **ID'** | **SC-$\delta$** | **Yes** | 0.21 | 0.07 |
| **PA** | **ER** | **Yes** | 0.39 | 0.27 |
| **PA** | **ID'** | **Yes** | 0.16 | 0.07 |
| **PA** | **SC-$\delta$** | **Yes** | 0.53 | 0.19 |
| **SC-$\gamma$** | **SC-$\alpha$** | **Yes** | 0.57 | 0.22 |
| **SC-$\delta$** | **SC-$\alpha$** | **Yes** | 0.1 | 0.05 |
| **SC-$\delta$** | **SC-$\gamma$** | **Yes** | 0.09 | 0.06 |
| **SC-$\delta$** | **PA** | **Yes** | 0.18 | 0.1 |
| **SC-$\delta$** | **ID'** | **Yes** | 0.11 | 0.06 |
| **SC-$\delta$** | **d** | **Yes** | 0.12 | 0.06 |
| PA | d | Yes (once) | 0.59 | 0.36 |
| SC-$\alpha$ | SC-$\delta$ | Yes (once) | 0.63 | 0.27 |
| SC-$\gamma$ | SC-$\delta$ | Yes (once) | 0.52 | 0.21 |
| SC-$\delta$ | SC-$\beta$ | Yes (once) | 0.6 | 0.16 |
| d | PA | No | 0.98 | 0.68 |
| d | SC-$\delta$ | No | 1.14 | 0.31 |
| ID | PA | No | 0.71 | 0.32 |
| ID | SC-$\delta$ | No | 2.75 | 0.73 |
| PA | PA - d | No | 0.94 | 0.59 |
| PA | ID | No | 0.72 | 0.36 |
| PA | SC-$\alpha$ | No | 0.73 | 0.29 |
| PA | SC-$\beta$ | No | 1.22 | 0.56 |
| PA - d | PA | No | 0.73 | 0.32 |
| SC-$\alpha$ | PA | No | 0.84 | 0.37 |
| SC-$\alpha$ | SC-$\beta$ | No | 1.06 | 0.33 |
| SC-$\alpha$ | SC-$\gamma$ | No | 3.19 | 0.91 |
| SC-$\beta$ | PA | No | 1.21 | 0.47 |
| SC-$\beta$ | SC-$\alpha$ | No | 1.11 | 0.43 |
| SC-$\beta$ | SC-$\gamma$ | No | 0.88 | 0.38 |
| SC-$\beta$ | SC-$\delta$ | No | 1.86 | 0.56 |
| SC-$\gamma$ | SC-$\beta$ | No | 1.03 | 0.53 |
| SC-$\delta$ | ID | No | 3.82 | 0.99 |

Table 2.3: Recovery of different kinds of $\Delta_{0.5}(Gen_1, Gen_2)$ generators. The mean fitness and dissimilarity values and whether the generator (with/without bloat) was recovered more than once (yes, in bold) out of the 5 runs for each pair of $Gen_1$ and $Gen_2$ is tabulated.

# Chapter 3

# Experiments with Empirical Networks

## 3.1 Subway Networks

We have with us the data of subway networks of many cities from around their time of inception till about 2009 from Roth C et al., 2012 - A long-time limit for world subway networks. What are the kinds of generators will we get by running `synth` on it and moreover can we use this dataset to validate the generators produced by Synth?

### 3.1.1 Quirks of the dataset

We know that adding nodes as an action is not supported by `synth` and we initialise all the involved nodes beforehand allowing only edge addition. However, in the case of the Barabasi-Albert model where edge addition and node addition happen while maintaining a constant average degree and `synth` successfully retrieves the PA generator consistently. We analyse the number of nodes and edges and the average degree of the subway networks of 14 major cities around the world since their inception till 2009. We see that the average degree remains more or less constant after an initial transient (which too exists only in a few cases).

So we are tempted to go ahead by just initialising all the nodes at the beginning. The largest network is 433 nodes and 475 edges, so getting generators that consistently generate networks of the best possible fitness will be rarer because 475 out of the $\binom{433}{2}$ possible edges is just 0.5% of the total set of edges that `synth` needs to pick probabilistically and consistently.

### 3.1.2 Simulations

With these quirks in mind, we proceeded to put them through `synth`. We ran the latest available networks (2009 or 2010) of Beijing, Berlin, Hong Kong, New York, Osaka and Shanghai. We ran each network 15 times with only mutation strategies, 15 times with mutation and recombination strategies. The average values of the generator with the best solution obtained as well as that of the number of generations in which the best solution was arrived at is in Fig. 3.1.

From the results, we can see that recombination makes getting to the final solution consistently quicker. However, the best fitness values that are obtained from the process are still better for `synth` without recombination. Given our experience with the two snapshot `synth` we know that using a better population of generators to mutate and recombine from can lead to improvements both in fitness values and the number of generations needed to get to the final solution. There we had 4 - best solution, best fitness of snapshot 1, best fitness of snapshot 2, and best fitness of both snapshots. Here, we have two of our three solutions to be very similar - best fitness and best mean dissimilarity, the generators maximising for these might also not be too different from one another because one solution optimises for maximum and the other for the mean value of the fitness of a solution that iteratively improves the worst of its attributes.

Therefore, if we expand the generator population with the quantities relevant to the ideal final solution we want while at the same time maintaining considerable difference amongst the pre-existing solutions that the generators are optimising for, we can hope to improve the fitness of the solutions gotten. In effect, rendering using recombination unambiguously better than using `synth` without it. So, what are some possible quantities that the generators should optimise for?

We observed that the progress in finding better solutions usually was stuck because of the lack of a good solution that had better similarity of the undirected distance attribute. We introduced the fourth solution as the one that optimised for the best undirected distance and used this population to run the cities 15 times through this updated recombination strategy and its results are also seen in Fig. 3.1. We can see that it improves in both fitness as well as the number of generations taken to obtain the best solution. This shows that while recombination can make it quicker to find solutions, it has to be combined with a good population of diverse solutions that will lead to better fitting solutions.

Another vital point to be made is that, none of the 6 subway networks got a nontrivial solution that repeated *i.e.* of the 15 runs for each city (using each strategy), all of them were different to one another. This is possibly because of the small size of the networks. Other possible contributing factors include the lack of information of the subway networks' spatial embedding given to the model which could be modified by including a new variable containing this information added to the vocabulary. It could also be because the topology of the network could've been especially hard, like in the case of lattices, where the inherent stochasticity makes it harder for the algorithm to give out exactly replicable lattice structures.

## 3.2 Street Networks

Therefore, using OSMnx [41] - a Python module for downloading geographical data from the OpenStreetMap API, we proceeded to model and analyze street networks of cities. The nodes are street junctions and the edges are the streets connecting two junctions. Given below is one 1.5 Km$^2$ region of each city's street network. Presented also are the streets of

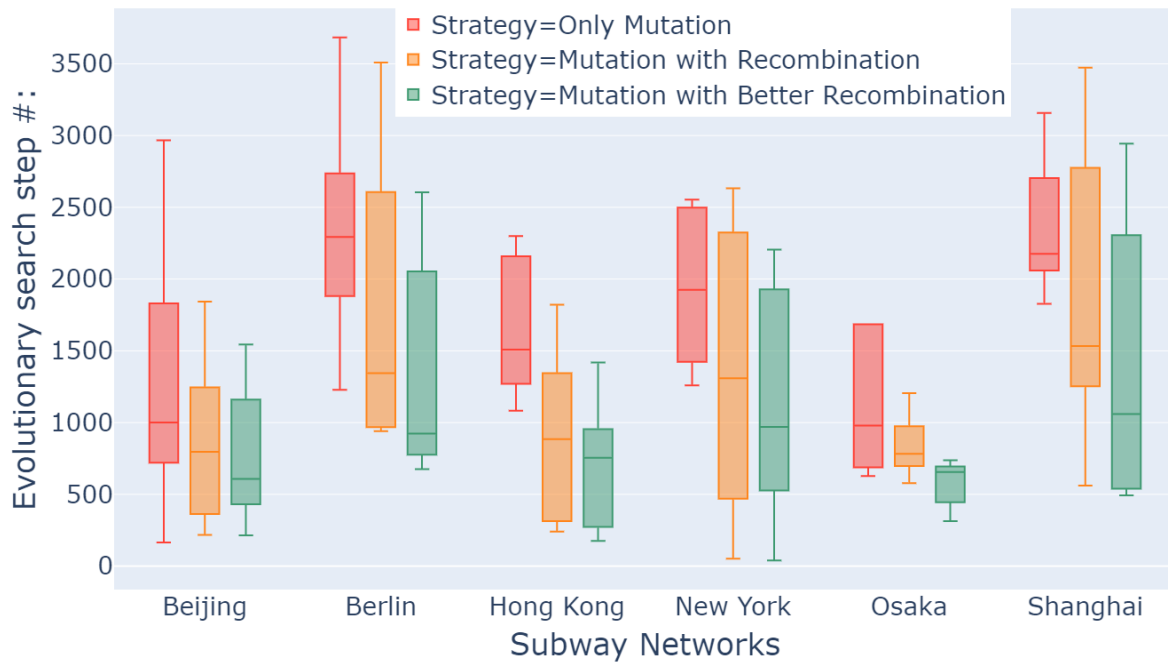

(a) The evolutionary search step number at which the best solution is reached is plotted for the subway networks of 6 cities using three evolutionary search strategies.

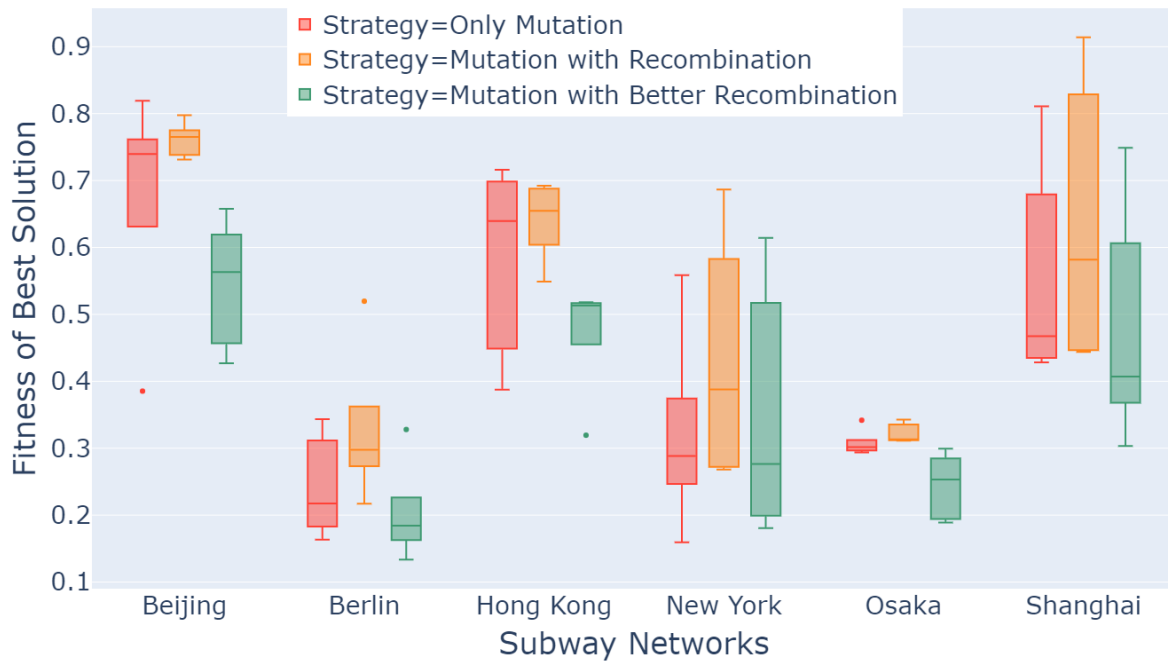

(b) The Fitness of best solution obtained using three evolutionary search strategies is plotted for the subway networks of 6 cities.

Figure 3.1: We compare three strategies - 1) Mutation, 2) Mutation and Recombination with population of solutions {best solution, best fitting solution}, 3) Mutation and Recombination with population of solutions {best solution, best fitting solution, solution with lowest mean dissimilarity, and solution with lowest undirected distance}. Each subway network was run through `synth` 15 times using each strategy.

Istanbul and Sao Paulo for comparison. We start by looking at 4 cities with organised streets, with the intuition that it might be easier for `synth` to find a structure for a network with less intrinsic randomness. For instance, compare the street maps of 1. Chandigarh and 2. Sao Paulo.

We can see that the first map is more ordered than the second, in order to quantify this we plot the polar histogram of the four cities we are considering: Chandigarh, Chicago, Dusseldorf and Manhattan as well as that of Istanbul and Sao Paulo for comparison.

In the polar histogram of every city, each bar's direction represents the compass bearings of the streets (in that histogram bin) and its length represents the relative frequency of streets with those bearings. Chicago is well aligned with all its streets in the N-S axis and the E-W axis. Manhattan is also well aligned but the axes are tilted along NW-SE and NE-SW. Dusseldorf has small bins which are not precisely aligned along their largest bins indicating some streets that are not in line with the rest. Compare it to Sao Paulo and Istanbul where you can see considerable number of streets flowing in all directions. Running the entire city network will be computationally infeasible so we have resorted to using three non-overlapping regions for each city. Each of the three samples has around 1200 nodes (each sample covers 1.5 Km$^2$). Their average degrees were coincidentally very similar - [Chandigarh, Chicago, Dusseldorf, Manhattan] : [2.57,2.51,2.34,2.38], implying around 3000 edges for each network. We wanted to check if the networks we chose were similar to each other, we computed the mean dissimilarity between each pair of cities and plotted it in Fig. 3.4. We see that all 4 cities are fairly similar to each other. This gives an expectation that all 4 cities might have related generative mechanisms.

We then proceeded to run each of these 12 (4 cities x 3 samples/city) networks for 3 runs each through `synth` (for a total of 36 runs). The results are listed in table 3.1

Except for Manhattan, it was found that each of the three sub-networks for each city had a common solution in at least one of their three runs. For instance, the three sampled

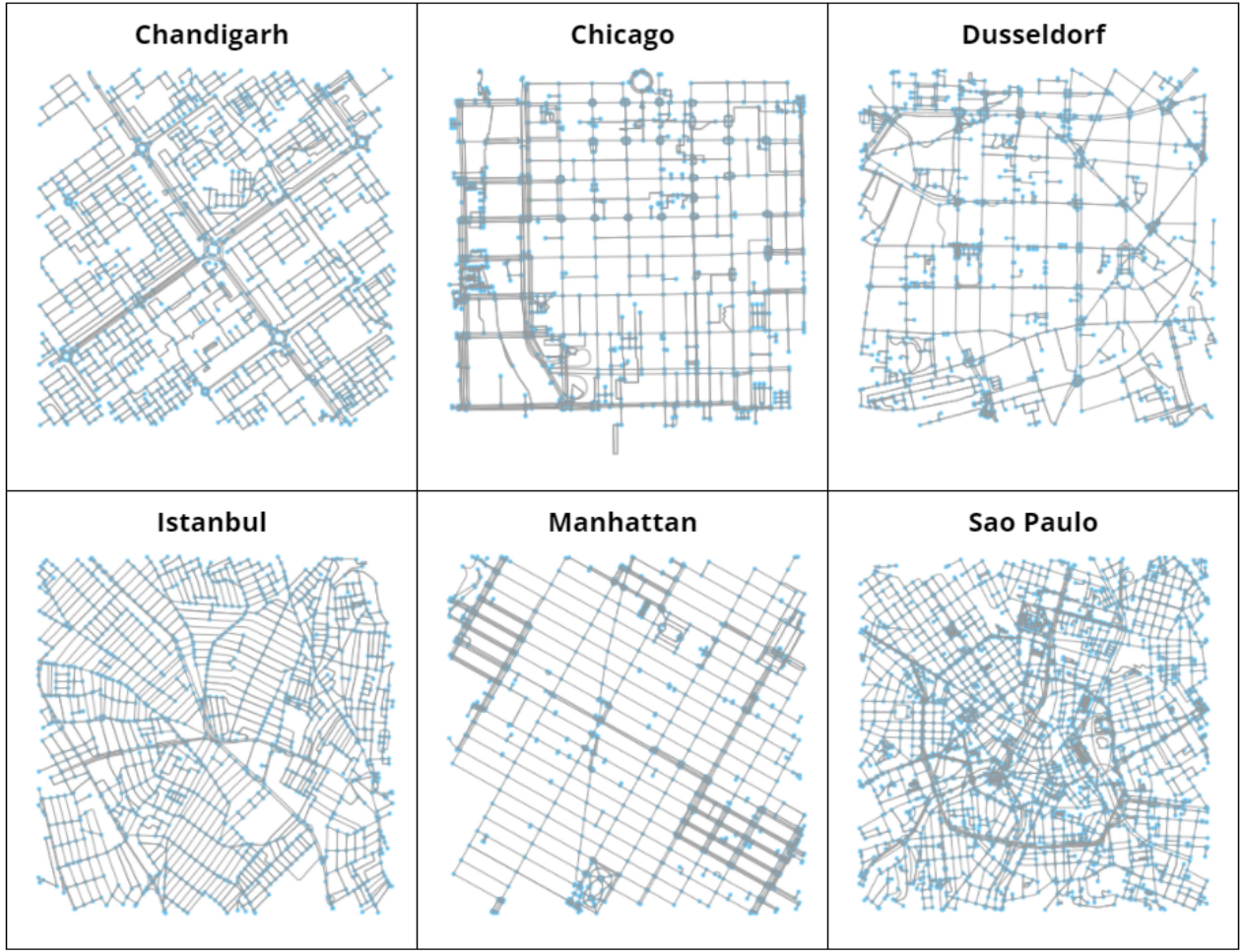

Figure 3.2: Street network of a 1.5 Km$^2$ region chosen for 6 cities. The nodes represent junctions and the edges represent streets connecting two junctions. These are simple, undirected networks without self-loops and multiple edges. The geospatial data was obtained from the OpenStreetMap API and the data was modelled and visualized as a network using a Python package - OSMnx [41]

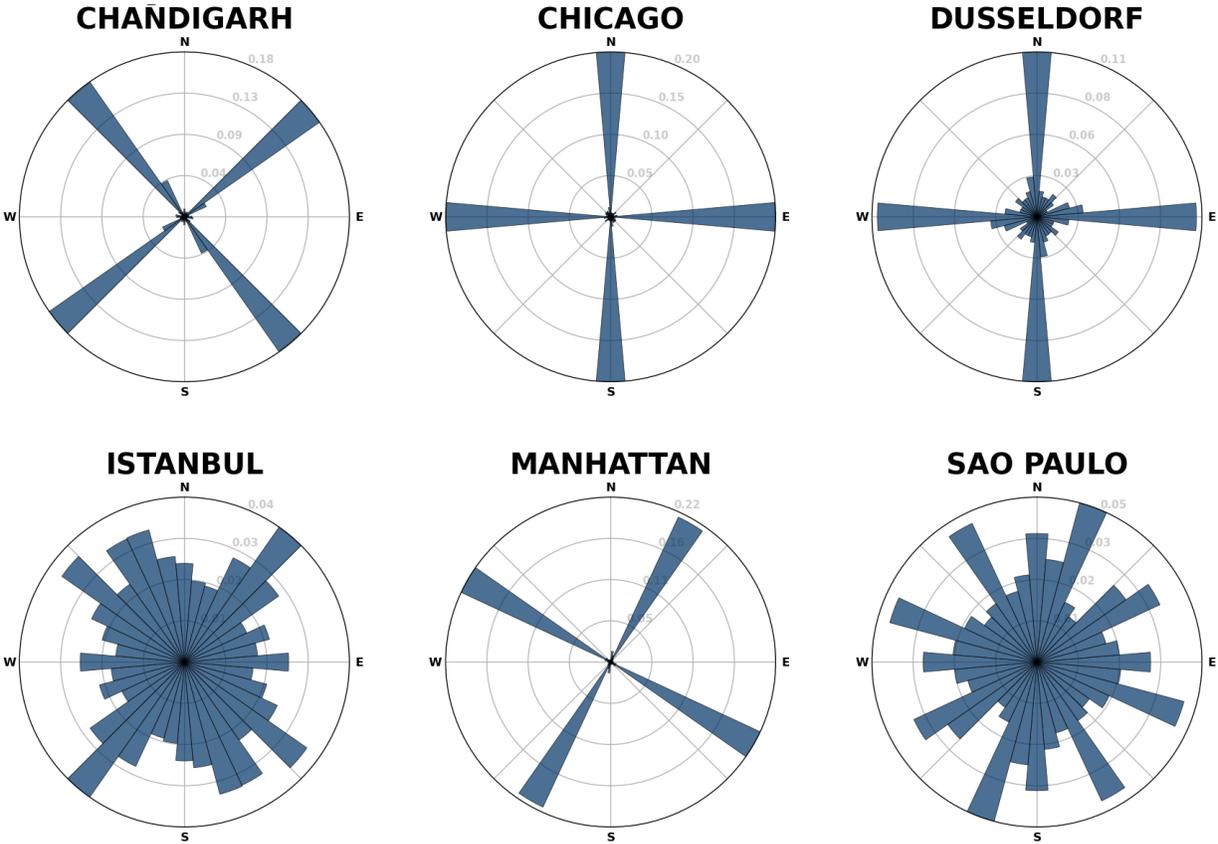

Figure 3.3: Plotted are the polar histograms of the street network of the 6 entire cities (municipality region). The length of each bar shows the relative frequency of streets along a certain directional axis, whereas the direction represents the compass bearings of the streets (in that histogram bin). An entropic measure used in Ref. [42] is used to calculate the disorderedness of each city's streets.

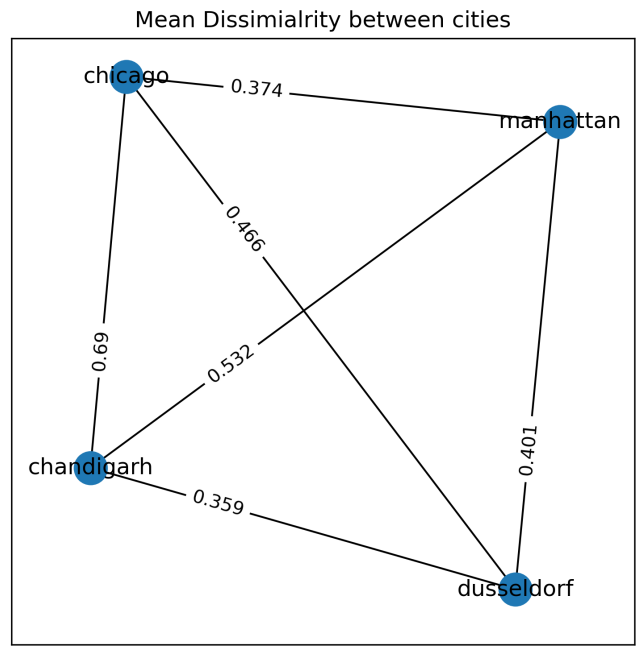

r

Figure 3.4: The pairwise mean dissimilarity between cities is calculated. All three regions of the city 1 are compared pair-wise with city 2 (and city 2 with city 1) and their values are averaged over.

| | Robust solutions of: | Dusseldorf, Chandigarh | Dusseldorf | Chicago, Chandigarh | Chicago, Chandigarh | Best values from |
|---|---|---|---|---|---|---|
| | With Generator: | $d(i,j)^{(0.82-k_i-k_j)}$ | $d(i,j)^{-2 \cdot k_i}$ | $2.5 - k_i \cdot \log(d(i,j))$ | $0.15^{(k_i^{d}(i,j))}$ | city's best generator: |
| Chandigarh | Fitness | **0.535** | 0.599 | 0.586 | 0.602 | 0.518 |
| | Dissimilarity | 0.436 | 0.493 | 0.526 | 0.532 | 0.5 |
| Chicago | Fitness | **0.610** | 0.712 | 0.603 | 0.627 | 0.386 |
| | Dissimilarity | 0.452 | 0.531 | 0.515 | 0.549 | 0.359 |
| Dusseldorf | Fitness | **0.467** | 0.531 | 0.63 | 0.619 | 0.5 |
| | Dissimilarity | 0.418 | 0.492 | 0.512 | 0.533 | 0.468 |
| Manhattan | Fitness | **0.844** | 0.799 | 0.92 | 0.965 | 0.711 |
| | Dissimilarity | 0.57 | 0.623 | 0.59 | 0.599 | 0.655 |

Table 3.1: In the top row are the cities with the best generators corresponding to them below. Each generator corresponding to a column is use to compare the dissimilarity and fitness between each of the four cities and the generator. The first column produces the best fitness and dissimilarities for all four cities compared to the other three columns. In the last column is the best fitness/dissimilarity value obtained for each city while those city networks were run through `synth`

networks for Chandigarh had the common solution: $2.5 - k_i \log d(i,j)$. While Chicago had a common solution: $(0.15^{k_i})^{d(i,j)}$. It gets interesting because Chandigarh's best generator was retrieved by one of the runs for Chicago and vice versa. This was observed again when Dusseldorf and Chandigarh shared the solution: $d(i,j)^{0.82-k_i-k_j}$. Are the solutions for one of the cities capable of explaining the others as well? To answer this we took all the robust solutions of the cities we obtained from 36 runs. A robust solution for a given city here means that at least two of three sampled networks had a common solution. We have four unique robust generators. Apart from the aforementioned three, we also have $d(i,j)^{-2 \cdot k_j}$. We check the fitness of these 4 solutions against all four cities. Each generator's fitness against the three networks for a given city was calculated. This was repeated ten times and the Mean Fitness and the Mean Dissimilarity (Overall) was calculated. The results of these are given below.

We see that the $d(i,j)^{0.82-k_i-k_j}$ solution explains 10/12 regions, the remaining two belong to New York. If we look at the functional form of the generator, the weight distribution curve does not change much for values of $k_i + k_j > 2$ and very quickly tends to zero for values of distance greater than 2. This can be interpreted as nodes that are one or two steps away from being connected tend to form an edge, and since this is true for all pairs of nodes, we get a very homogeneous degree profile explaining the uniform pattern of the network.

The mean fitness values for the robust generators are mostly around 0.5 (except Manhattan) while this may initially seem like there's a pressing need for better generators that can explain all cities better (which is still a possibility), these generators are actually doing pretty well for themselves. Given in the last column for reference are the best fitness and dissimilarity values gotten for each city during each of their 9 runs. We can see that the best possible fitness values `synth` could find for each of the cities through generators that are not necessarily robust happen to be not too far away from the fitness values of the robust generators. This seems to indicate that `synth` is retrieving solutions of good fitness values and while the generator expression corresponding to those values vary there still exist robust solutions amidst them

Amongst these 4 cities, Manhattan is the weirdest. It looks very structured and similar to the rest but except in 2 runs the remaining 7 had mean fitness values ¿0.9, indicating that `synth` was not able to find good solutions. The other rationale is that the ER generator explains the city well but we can visibly see that it might not be the best resolution. The highly ordered lattice like structure of the network and the stochastic nature of the generator mechanism of `synth` no doubt have a part to play. From this analysis, we wish to point out that `synth` provides promise and is a viable candidate for figuring out city structures.

## 3.3 Semantic Cooccurrence Networks

The corpus of scientific literature is growing at a fast pace. The number of papers every month grows exponentially. Especially, the field of Artificial Intelligence (AI) and Machine Learning (ML) has a doubling rate of roughly 23 months ¡insert reference of IARAI graph¿. The Institute of Advanced Research on Artificial Intelligence (IARAI) have created a semantic co-occurrence network capturing the content of scientific literature in AI since 1994. Each of the 64,000 nodes in the network represents an AI concept. Edges between nodes are formed when the two corresponding concepts are studied together in a research paper. This unique dataset offers different kinds of explorations on its own. We are interested in looking at dynamic networks using `synth` to see if it can say anything about the generative mechanism of such networks. Therefore, we decided to look at single concepts and their corresponding ego-centered networks[43] and their evolution over time and to look at the trends longitudinally and whether there are similarities. An ego-centered network (of radius 1) is one where we fix a node (the ego) and look at all its neighbours and the edges between this subset of nodes We created the ego-centered networks of 10 nodes at two time points - at 2014 and at 2017. We ensured that the networks of the 10 nodes had no common nodes in an effort to contain the bias introduced due to snowball sampling [43]. We also ensured that the sizes of these ego-centered networks were not too large owing to computational constraints. These 10 networks at 2 time points were run through `synth` and the generators recovered are tabulated in table 3.2.

We see that all the 2017 snapshots have best solutions with fitness better than their 2014 counterparts. Taken along with the fact that the ratio of edges in the network to the total possible edge set ($R$) is also lower for the 2017 snapshots, it might be tempting to contest the point previously made that networks with lower values of $R$ would have it harder to find good solutions. However, it has to also be noted that the network sizes are significantly larger for the 2017 snapshots, leading to the possible explanation that more distinctive generators are arrived for these larger networks because there are more number of edges that the generator expression has to correctly predict to satisfy the fitness condition.

Now with regard to the generators themselves, a lot of the 2014 snapshots have ID or ER generators, the ID generator could be because of our choice to look at ego-centered networks resulting in the unavoidable bias introduced by the central node and all nodes in consideration are chosen because they have an edge with it, so all these nodes are going to be chosen at least once. So the edges could've been added randomly but since all nodes in the network are involved with the central node, the ID generator or the ER generator could work as the ordering doesn't matter because at the end of the network creation, we are going to get all nodes connected to the ego node. This also supports our hypothesis that the network size plays a crucial role in giving out meaningful and distinct generators because the same study for the 2017 networks resulted in generators with more variety.

Some of the generators are in the form of the $\Delta$ function implying a time component, a point after which the character of the network has changed. If we look at all such generators in the 2017 cohort, we see that the first generator is in fact the 2014 generator (ID or ER) and the point of switching generators happens after the 2014 snapshot. We are able to say this because all the points of switching are greater than the $\frac{R_{2014}}{R_{2017}}$, indicating that a distinctive character has developed between these two points. Most of the generators explaining the post-2014 parts of the network seem to be form PA and PA' families (after simplification or removing bloat).

It would be an interesting study to compare networks at a later time (say 2020) and look at whether the characteristic generators are carried over there too or maybe there's a further change in dynamics. This would help to get an extended temporal perspective and provide to be an opportunity to improve/test the robustness of the 2017 generators.

| Node ID | Year | N | E | $R = \frac{E}{\binom{N}{2}}$ | $\frac{R_{2014}}{R_{2017}}$ | Best Fitness | Generator |
|---|---|---|---|---|---|---|---|
| 5697 | 2014 | 36 | 375 | 0.595 | 0.203 | 0.71 | ID' |
| 5697 | 2017 | 96 | 1847 | 0.405 | | 0.56 | ID' |
| 7296 | 2014 | 41 | 363 | 0.443 | 0.448 | 0.97 | ER |
| 7296 | 2017 | 77 | 811 | 0.277 | | 0.61 | $\Delta_{0.5}(ID, e^{PA})$ |
| 14446 | 2014 | 21 | 210 | 1 | 0.417 | 0 | ER (or any generator) |
| 14446 | 2017 | 38 | 503 | 0.716 | | 0.45 | $\Delta_{0.79}(ID, ER)$ |
| 14628 | 2014 | 26 | 325 | 1 | 0.258 | 0 | ER (or any generator) |
| 14628 | 2017 | 73 | 1259 | 0.479 | | 0.45 | $\Delta_{0.62}(ER, PA)$ |
| 48598 | 2014 | 66 | 750 | 0.35 | 0.295 | 0.71 | $PA^2$ |
| 48598 | 2017 | 139 | 2544 | 0.265 | | 0.23 | $PA^2$ |
| 56970 | 2014 | 43 | 449 | 0.497 | 0.531 | 0.51 | $PA^{2^{ID}}$ |
| 56970 | 2017 | 63 | 845 | 0.433 | | 0.44 | $PA^2$ |
| 3134 | 2014 | 40 | 509 | 0.653 | 0.160 | 0.74 | ID |
| 3134 | 2017 | 130 | 3187 | 0.38 | | 0.25 | $\Delta_{0.62}(ID, PA)$ |
| 3426 | 2014 | 42 | 557 | 0.647 | 0.142 | 0.82 | ID |
| 3426 | 2017 | 142 | 3920 | 0.392 | | 0.48 | $ID^5$ |
| 10748 | 2014 | 53 | 1068 | 0.775 | 0.410 | 0.35 | $(> i\,9\,ID\,ER)$ |
| 10748 | 2017 | 92 | 2604 | 0.622 | | 0.33 | $\Delta_{0.61}(i^i, i^j)$ |
| 19215 | 2014 | 42 | 561 | 0.652 | 0.298 | 0.73 | ID |
| 19215 | 2017 | 96 | 1880 | 0.412 | | 0.45 | $\frac{PA}{ID}$ |
| 39065 | 2014 | 112 | 1266 | 0.204 | 0.382 | 0.71 | $\Delta_{0.61}(ER, ID')$ |
| 39065 | 2017 | 181 | 3311 | 0.203 | | 0.38 | $\frac{PA}{ID}$ |

Table 3.2: 10 ego-centered networks were created from the semantic co-occurrence network of scientific literature in Machine Learning such that no overlap existed between the networks. Each network at 2014 and at 2017 was run through `synth` 5 times. The best generator for each network is displayed in the last column.

# Chapter 4

# Conclusions

## 4.1 Tying it all up

In this thesis we described and appaised and reproduced the results obtained using the symbolic regression framework (`synth`) proposed in Ref. [9]. We then explored the limitations of the framework, one of which was to try to overcome the lack of a network growth mechanism. We showed how it was incongruous to incorporate the conventional growth mechanism of node addition into the framework. Another limitation was the lack of sufficient semantics to capture dynamic networks. For which we implemented extensions to accommodate dynamic networks into `synth`. We have expanded the vocabulary to be able to study networks that are created in phases by multiple generators using edge ratio and the $\Delta$ function. We were able to see the existence of distinctive characteristics of the $\Delta$ function for many (not all) pairs of generators. Another constraint we tried to reduce was to make the framework run more quickly for which we used the random walk heuristic distance measure to remove bottlenecks in runtime of the code. A meaningful improvement was however when we introduced recombination of solutions to aid in `synth`'s ability to recover the $\Delta$ generator (and the rest of the gamut) consistently.

With the improved framework, we ventured into gathering a few encouraging results by using the framework on subway, street and semantic networks. We saw that in all three sets of networks we used, we got solutions with good fitness values that could reconstruct the networks stochastically. The city and semantic networks gave out generators that were interpretable and that could explain the network formation process. Running similarly orientated city networks with comparable dissimilarity measures through `synth` gave us robust generators that explained the cities as well as one nearly universal generator for almost all of the city regions considered, making a case for the explanatory power of the framework's solutions. The semantic co-occurrence networks gave out many $\Delta$ functions with recurring constituent generators ($ID$, $PA$, $PA'$) indicating a change in the nature of the networks generated from sequential identifiers because of the ego-centered nature of the networks and moving to distinctive generators.

While the subway networks were too small for `synth` to find rules for, we were able to use the size to our advantage and test the effects of recombination on improving the the evolutionary steps it takes to obtain the best solution as well as the fitness of the best solution found. We were able to see that recombination when used with a diverse population of solutions does indeed help in getting to quicker and better solutions. While the lack of consistent generators in the case of the subway networks could be improved by introducing topological information of the nodes and edges and accommodating them into the vocabulary, we used these empirical datasets to showcase the possibilities of the framework in producing interpretable growth processes. For more specific datasets that need to be analyzed, scientists can use this framework and modify the semantics according to their domain and use case to understand the underlying process to aid in their modeling ventures.

## 4.2 Future Course Of Work

There are many ways in which the framework could be improved and taken further. The existence of independent variables in the generator vocabulary results in a multidimensional search space of generators. It is an interesting theoretical problem to find out the recovery rates of solutions of `synth` across the regions to spot the resolution limits of not just this framework but hopefully the generators themselves. Such an exploration will then involve optimising the code to scale for many runs of network generator recovery as well as recovery of larger networks since sparse empirical networks might not have a consistently robust signature. Another source of help might be the multi-snapshot version of the framework where we are able to causally connect the information provided and use it to bias the framework towards better solutions quicker.

Exploring specific use cases always have potential. For instance, introducing semantic similarity variables (like Jaccard's coefficient) and try to see if `synth` can capture and explain underlying group structure in social networks by using this more advanced vocabulary.

In our framework we had allowed for only addition of edges. However, deletion of edges and if possible using another method, a provision to add/delete nodes are also avenues of improvement. A hotbed of questions awaits, in my humble opinion, regarding the theoretical understanding of the properties of networks created in stages - using multiple generators and how that affects network structure. The $\Delta$ function introduced to study this is hopefully just the beginning.

# Appendices

# Appendix A

# Methods and Techniques

This chapter contains details about the implementation of the `synth` in its original form and some of which will be carried over to the extension we intend to make with this framework. Wherever we choose to improve a technique we explain.

## A.1 Anti-bloat tolerance

The anti-bloat tolerance ratio $b_r$ is introduced to keep the generators from becoming larger in program length and becoming bloated in size [31, 44]. If the weight distribution arrived at is the same then the algorithm is set to prefer the generator that is less complex (the one with smaller program length).

Like the sampling ratio, $b_r$ also introduces a trade-off between generator length and the accuracy of the results. If the $b_r$ is too large and lenient, then we are essentially optimising for generators with better fitness with less regard to the computational costs that are involved every time to create a network from a large generator.

## A.2 Dissimilarity Metrics

For the degree and PageRank centralities where there is a well defined notion of distance between the bins of the distributions, we use the *Earth Mover's Distance*. Intuitively, if the distributions over a specific region are viewed as two different ways of piling up a specific amount of dirt (earth) over the specified region, the EMD is the lowest cost of changing one pile into the other, where the cost is supposed to be the amount of dirt transported times the distance travelled.

For the more complicated distance distributions and triadic census, we calculate the difference between the counts of the category in a quantity (say a category - *003* triad among the quantities - 16 triads) for the network generated synthetically using the given generator and the target input network and divide this with the counts of the *003* triad in the synthetic category. We sum over all such categories and use that as the dissimilarity. Suppose there are $K$ categories and we denote the counts of the $i$th category for synthetic and target network as $c'_i$ and $c_i$, then the dissimilarity $d(c, c')$ is given as:

$$d(c, c') = \sum_{i=1}^{K} \frac{|c_i - c'_i|}{n_0(c'_i)}$$

where $n_0(c'_i) = c'_i$ for all nonzero values and 1 otherwise. This function prevents division by zero and unnecessary singularities.

## A.3 Generator initialisation

In section 1.3, the evolutionary search process was started by initialising a randomly created generator. There are two strategies employed to create a randomised tree - *fixed depth* and *grow*. Irrespective of the strategy, for each generator tree a tree minimum depth variable $D$ is assigned a value $D \in [D_{min}, D_{max}] \mid D_{min}, D_{max} \in N$. The fixed depth strategy creates a balanced size tree, all nodes of the tree at the minimum depth value are terminal nodes while all nodes below the a tree depth of $D$ are parent nodes *i.e.* they are function nodes. For the former, variables or constants are chosen with equal probability. All the nodes that have a tree depth less than $D$ are picked randomly from the list of available function nodes ¡insert table for functions¿. Note that some function nodes (or operators) have 1/2/3/4 terminal nodes (operands), so the number of nodes in a depth level is dictated by the number of operands demanded by the operator in the previous depth of the tree. The grow strategy too assigns all the nodes below the minimum depth to be function nodes chosen with equal probability. However, for nodes at the depth level $D$ there is a probability ($p_{terminal}$) with which a terminal node *i.e.* a variable/constant is chosen.

The choice of either a variable or a constant is equiprobable. In case of a variable, one of the entries from table 1.2 are chosen with equal probability. In case of a constant, 0 is chosen with 0.1 probability, an integer in the uniform distribution $\{0, ..., 9\}$ is chosen with 0.4 probability or a real number from uniform distribution $[0,1]$ is picked with 0.5 probability.

These two strategies are standard techniques adapted from [9, 45] genetic programming literature. The parameters used are listed in table **??**, we use this approach for creating variety in the generator tree. The actual values are carried over from [9, 17], it's probable that the precise values and parameterization have minimal effect on the search algorithm's results and we need not worry about optimising them.

The process of mutation involves choosing a sub-tree in the generator tree and replacing it with another sub-tree. This replacement is obtained by first randomly generating the equation tree as mentioned above, and choosing one node from this tree with uniform probability and picking the sub-tree with the chosen node as its root. This is known as headless chicken mutation [46].

Recombination involves picking two trees from the pool of solutions available. A node from parent 1 and a node from parent 2 is chosen with uniform probability. The sub-tree

| Parameter | 'Value |
|---|---|
| Sampling ratio $s_r$ | 0.0006 |
| Anti-bloat tolerance $b_r$ | 0.10 |
| Bins for distributions | 100 |
| $D_{min}$ | 2 |
| $D_{max}$ | 5 |
| $p_{terminal}$ | 0.4 |
| "Infinite" distance value for disconnected nodes | 10 |
| Stable evolutionary steps; stop condition | 1000 |

Table A.1: Parameter values used in our runs.

with the chosen root node from parent 2 replaces the sub-tree with chosen root node from parent 1.

## A.4 Random Walk Heuristic Distance

A very resource intensive part of the algorithm is the function that calculates the (simple/directed/reverse) distance between two nodes. We employ the industry standard Dijkstra algorithm $[\mathcal{O}((V+E) \cdot \log(V)]$ to get the distance [22] and still have this component driving at a minimum of 23% of the total run-time of the algorithm. We consider an alternative heuristic measure that arrives at the distance measure approximately. This measure was introduced in Ref. [9] to deal with the same issue. While swapping this measure with the exact distance it tries to most importantly retain the essence of a "distance" value in the network even if the accuracy is reduced (not significantly, here) while notably reducing computational costs. In this regard, we use random walking (RW) to forage the network and iteratively arrive at better values of the distance between two nodes.

For a network of $N$ nodes, we create a $N \times N$ distance matrix $M = \{m_{ij}\}$ where $m_{ij}$ contains the heuristic distance from node $i$ to node $j$ (for an undirected network, $m_{ij} = m_{ji}$). We initialise this matrix with a value of 6 (we can arrive at this upper bound for the diameter of a network using Ref. [47], not to mention the popular wisdom of "six degrees of separation" [48]). We assign a RW for each node in the network. For every new edge added we make each RW carry out 5 steps (and after these 5 steps they are reset to their assigned origin node). During each of these 5 steps; if a walker visits a node ($j$, say) that is not its origin node ($i$), it checks the number of steps ($c (\leq 5)$) it has taken so far after being reset back to its starting node $i$. If this count is lower than the current distance ($c < m_{ij}$), then the distance matrix is updated with the count ($m_{ij} \leftarrow c$)

Our choice of using 5 RW steps per edge addition instead of 1 or 10 was a trade off between accuracy and computational costs. Using anything above 6 was redundant due to the average connectedness values of most synthetic and empirical networks we are dealing with [48, 47] while using 1 RW per edge addition was too low for all the walkers to sufficiently explore their neighbourhoods and in effect, the entire network. Therefore, we've plotted the fraction of node pairs whose distances have been correctly identified by the heuristic measure as a function of the fraction of edges that have been added in the network. We plot curves comparing 1 vs 5 RWs per edge added. We plot this comparison in Fig. A.1 using two different network sizes (200 nodes and 2000 edges as well as 300 nodes and 3000 edges) and we see that changing network size does not matter in the size ranges we are primarily working in. We see that there is a significant improvement if we use 5 RWs per edge addition and that roughly halfway in we have with us a very good approximation of a distance measure. Using this has led to a significant decrease ($\tilde{6}0\%$) in computational run-time of the part of the algorithm that calculates the distance.

We compared the networks created using the distance generator ($w_{ij} = d(i,j)$) with the exact measure and the RW heuristic distance (with 5 RWs per edge addition). We found that the dissimilarity between these two sets of networks to be relatively close (0.35). More importantly, the reliability to use these heuristic measures stem from the their ability to recover the generators that involve $d(i,j)$ consistently. In the inset of Fig. A.2, we present the recovery results of the distance generator using three methods, the first column is using the standard `synth` algorithm, the second column is using the RW heuristic Distance in place of the exact distance measure. The third column is the same as the second column but instead of just using mutation to create a better quality of solutions we also use recombination. In each of the three cases, we create 15 networks of size 200N∥2000E and find that all 15 are recovered in all three cases. The second and third methods have increasing fraction of bloat, as expected, in them [49], which when simplified gives us the distance generator. The main figure is a violin plot comparing how quickly or the number of generations (*i.e.* the number of evolutionary search steps) it takes to arrive at the final distance generator. The means have been plotted along with the extremities in each column. We see that the first two methods have roughly comparable mean values and probability distributions (shaded blue areas arrived at using kernel density estimation) on the y-axis. The third column, as expected, has a mean value that is significantly lower confirming our intuition of using recombination to arrive at the right solution quicker. Another interesting feature is the roughly uniform distribution of the values between the extremities for all three columns. This illustrates why the dynamic stopping condition we have used (*i.e.* the algorithm stops when no better solutions are found for 1000 generations). We can't risk stopping early for the possibility of losing out on the best solution because of the inherent randomness involved in mutation and recombination.

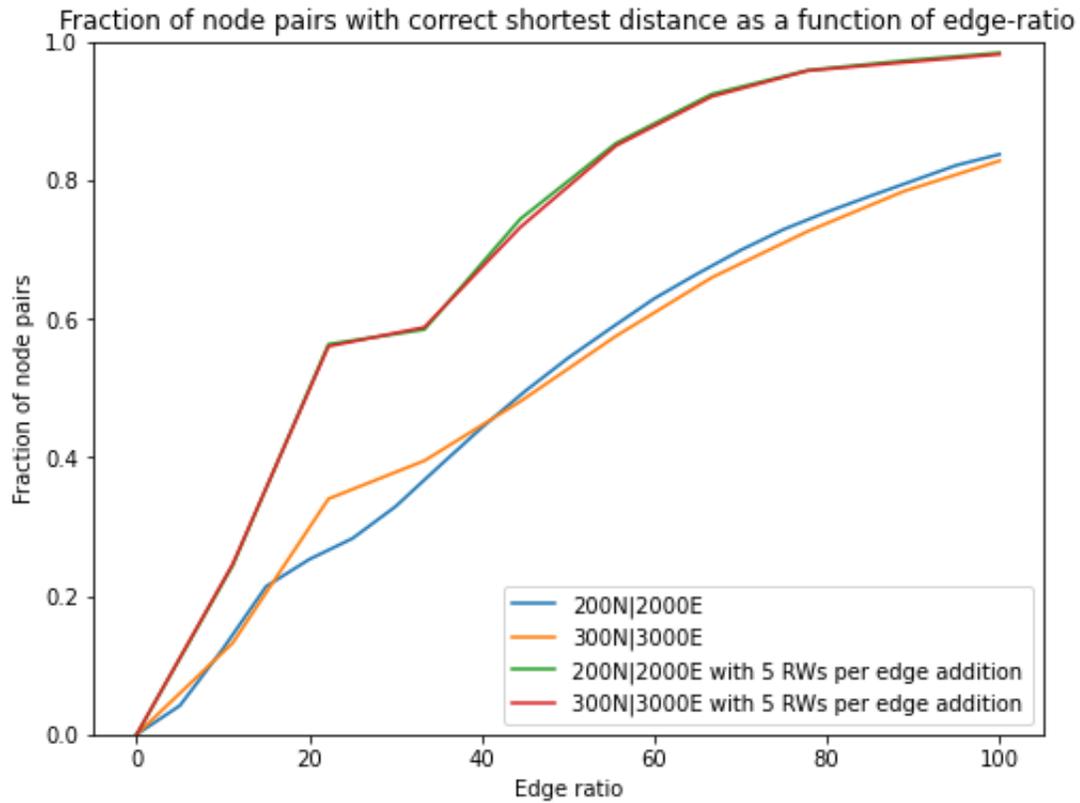

Figure A.1: The fraction of node pairs in the network whose shortest distance is correctly identified by random walk heuristic distance as more edges are added during the network creation process. Two strategies are compared - 1) for each new edge 1 random walk step is executed by each walker starting from its corresponding node in the network and 2) 5 random walks are executed per each edge addition step. We compare these two strategies using two different sizes of networks - 200N‖2000E and 300N‖3000E.

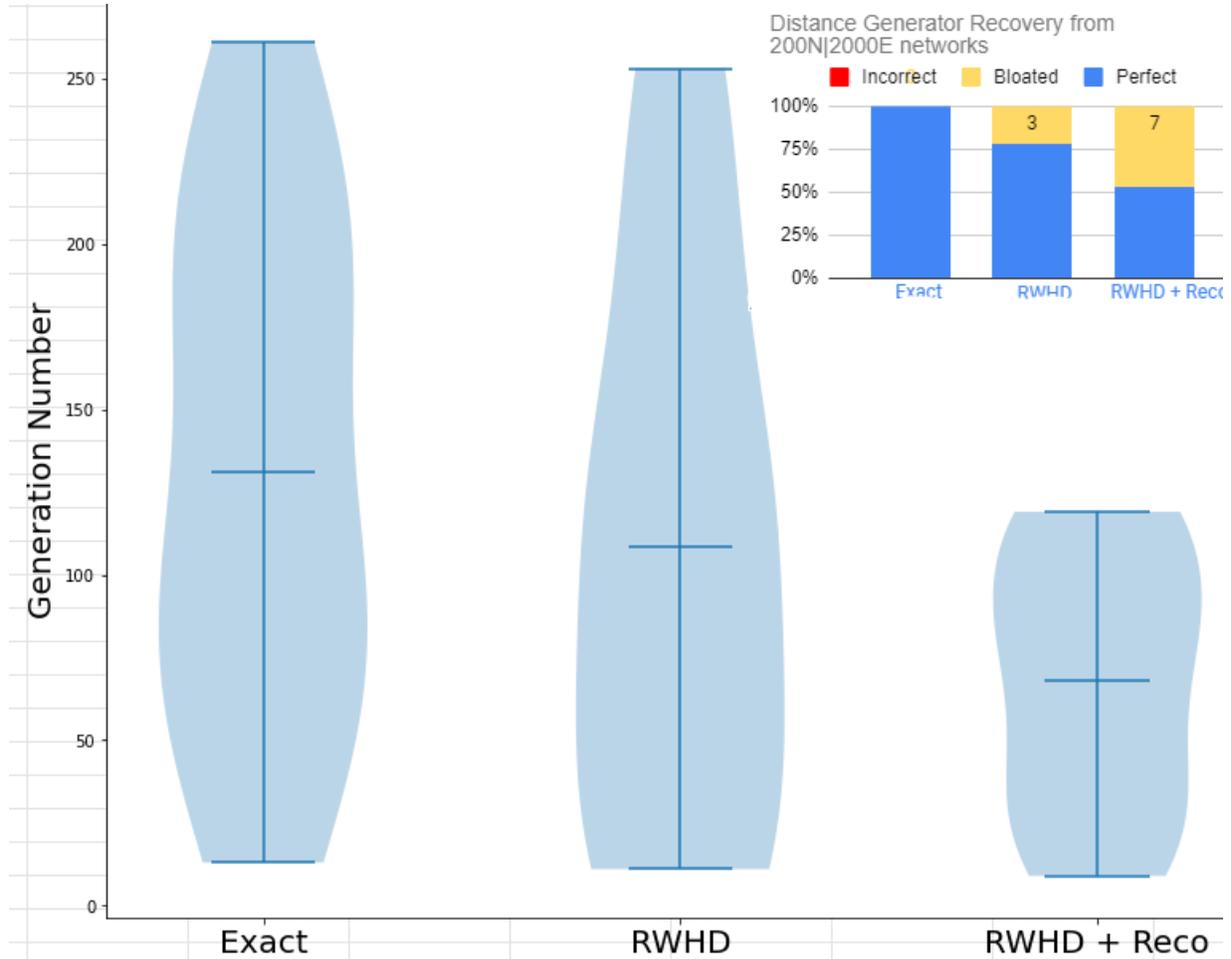

Figure A.2: The generation number at which the best solution is plotted for three strategies - 1) `synth` in its standard form *i.e.*, using exact distance measures and only mutation as search strategy 2) using the heuristic distance with 5 random walks per edge addition and only mutation as search strategy, 3) same as 2) but with recombination also as a search strategy. These recovery tests were done on 15 200N‖2000E networks generated using the $w_{ij} = d$ for each strategy. Plotted in the inset, is the percentage of recovery of the distance generator during the 15 runs for each of the three strategies.